# Predicting Space Groups of Double Perovskites by LLM with Dynamic Few-Shot Learning

Jongwon Park [1,†], Inhyo Lee[1,†], Junhyeong Lee[2], and Seunghwa Ryu[1,2,3 *]

**Affiliations**

[1]Department of Mechanical Engineering, Korea Advanced Institute of Science and Technology (KAIST), 291 Daehak-ro, Yuseong-gu, Daejeon 34141, Republic of Korea

[2]KAIST InnoCORE PRISM-AI Center, Korea Advanced Institute of Science and Technology (KAIST), Daejeon, 34141, Republic of Korea

[3]Department of AX, Korea Advanced Institute of Science and Technology (KAIST), 291 Daehak-ro, Yuseong-gu, Daejeon 34141, Republic of Korea

[*]Corresponding author: ryush@kaist.ac.kr

## Abstract

Double perovskites (DPs) offer broad compositional tunability, but predicting the space groups (SGs) of stable structures remains difficult because available datasets are often strongly imbalanced toward dominant SG classes. In this work, we introduce **Dy**namic and Diversity-enhanced few-shot retrieval and **R**ule-Guided **I**nference for **S**pace-Group Prediction (DyRIS), an LLM-agent-based framework that predicts ranked SG candidates from a given DP composition. DyRIS uses diversity-enhanced dynamic few-shot prompting to retrieve relevant in-context examples while limiting the dominance of frequently represented SGs. It further incorporates rule-guided inference based on *B/B′* cation ordering and quantitative indicators and rank the final Top-3 SG candidates. We evaluate DyRIS on 3,528 thermodynamically filtered DP entries and compare it with composition-based and descriptor-based baseline models. Considering the severe class imbalance in the dataset, we separately evaluate the prediction performance for underrepresented SG classes, referred to here as minor SGs. At a training-data ratio of 0.5, DyRIS achieves competitive overall accuracy while obtaining the best Overall Top-1 macro-F1 score and the best performance across all Minor-SG metrics. In particular, DyRIS improves Minor-SG Top-1 accuracy by 3.26 percentage points relative to CrabNet and achieves higher Minor-SG Top-3 accuracy than the strongest PyCaret-based baseline. Ablation studies show that diversity-enhanced retrieval, quantitative indicators, major-SG bias-control and B/B′ ordering information each contribute to prediction performance, while classifier- and ranker-based replacement experiments indicate that the final rule-guided inference step is not easily replaced by conventional machine learning models. Additional analyses reveal that DyRIS improves minor-SG prediction by integrating retrieval evidence, quantitative evidence, and crystallographic prior information, although its final ranking step can limit Top-1 performance in the high-data regime. These findings demonstrate the potential of combining retrieval-based LLM reasoning with crystallographic domain knowledge for SG prediction in imbalanced materials datasets.

## 1. Introduction

Double perovskites (DPs) generally represented by the chemical formula $A_2BB'X_6$, have attracted significant attention as next-generation functional materials.[1–3] This interest arises from their high compositional tunability, which enables targeted properties to be explored by combining different elements at the $A, B, B'$ and $X$ crystallographic sites.[4,5] Accordingly, DPs have been actively studied across a wide range of applications, including optoelectronics, thermoelectric, spintronics, and solar cells.[6–11]

The functional properties of DPs are governed not only by their composition but also by the crystal structures stabilized by that composition.[12] Even among DPs sharing the same composition, different space groups can produce markedly different properties—for example, symmetry changes have been reported to affect electron mobility and optical absorption, while octahedral tilting of $BX_6$ and $B'X_6$ units can modify carrier transport.[13,14] Since these properties depend on the stable structure a composition adopts,[15,16] identifying that structure is a prerequisite for the reliable design of DPs with targeted functional properties.

Identifying this stable structure, however, is far from straightforward. DPs span space groups from low to high symmetry,[16–18] as distortions—differences in each site ionic radii and octahedral tilting—readily alter lattice symmetry.[19–21] Consequently, experimentally determining stable crystal structures for a given DP composition requires substantial resources. As an alternative, high-throughput ab initio calculations have been proposed. However, this approach also requires prior information about the primitive unit cell to perform calculations. Moreover, for low-symmetry structures, the computational cost increases drastically as the number of atoms within the primitive unit cell grows.[22]

To overcome these limitations in computational cost, data-driven approaches have been introduced. Specifically, surrogate models such as CRYSPNet can predict possible space groups (SGs) from a given DP composition,[23] and generative models such as MatterGen, a diffusion-based model, have demonstrated new possibilities for materials discovery by directly generating crystal structures for a given composition.[24]

However, these data-driven approaches are inherently limited by their strong dependence on the training-data distribution.[25,26] In practice, DP structures contained in materials databases such as the Inorganic Crystal Structure Database (ICSD) and the Materials Project are biased toward specific structures such as *Fm-3m*.[23] Such dataset imbalance can degrade the prediction performance of purely data-driven models for minor SGs. Given the structural diversity of DPs, an approach that incorporates crystallographic domain knowledge beyond statistical patterns in the training data is needed.

To address these limitations, approaches leveraging large language models (LLMs) have recently gained attention in materials science.[27–30] LLMs, trained on large-scale literature data, implicitly encode a broad range of knowledge and rules related to materials science. In practice, LLMs have demonstrated strong performance in tasks ranging from materials-related question answering[31,32] to novel materials discovery.[33–35] Importantly, crystallographic rules and domain knowledge can be explicitly incorporated into the reasoning process through prompt engineering. This enables SG prediction to be guided not only by physically and crystallographically meaningful constraints but also by the broad materials-science knowledge embedded in LLMs, rather than solely by statistical correlations in training datasets. Therefore, an LLM-based framework integrated with domain knowledge offers a promising route for predicting stable SGs from DP compositions and exploring the structural diversity of DPs.

In this study, we propose **Dy**namic and Diversity-enhanced few-shot retrieval and **R**ule-Guided **I**nference for **S**pace-Group Prediction (DyRIS), an LLM-agent-based framework for predicting ranked SG candidates corresponding to stable DP structures from a given DP composition while addressing challenges arising from severe class imbalance. Although LLMs encode broad materials knowledge, their implicit knowledge alone is insufficient to reliably narrow plausible SG candidates. To address this issue, we introduce diversity-enhanced dynamic few-shot prompting, which retrieves in-context examples based on proximity in the embedding space to constrain the candidate space while ensuring balanced coverage across SGs. We further incorporate rule-guided inference that integrates *B/B′* cation ordering, quantitative indicators and major-SG bias-control, thereby improving logical consistency and prediction reliability. Our results demonstrate that DyRIS achieves comparable or superior overall performance and improves minor-SG prediction in low-data regimes. Even in the high-data regime, DyRIS remains competitive in several Top-1 metrics for minor SGs. These findings demonstrate the potential of combining in-context LLM reasoning with crystallographic domain knowledge as a promising complement to purely statistical data-driven approaches for materials prediction.

## 2. Methods

This section describes the construction and evaluation of the proposed DyRIS framework. We first explain the dataset preprocessing procedure, including stability filtering, SG-class filtering, and train/test splitting. We then describe the feature extraction process used to construct the embedding space. Next, we present the diversity-enhanced dynamic few-shot prompting strategy, in which in-context examples are dynamically retrieved for each query composition. Here, a query composition refers to the target DP composition for which SGs are predicted. We further describe the rule-guided inference procedure, in which *B/B′* cation ordering information, quantitative indicators and major-SG bias-control are used to refine and rank the final Top-3 SG candidates. Finally, we describe the experimental settings for baseline comparison, data-sensitivity analysis, and ablation studies. The LLM-agent component of DyRIS was implemented using GPT-5.4-mini as the backbone model, and the detailed LLM configuration is summarized in **Table 1**. The overall workflow of DyRIS is summarized in **Figure 1**.

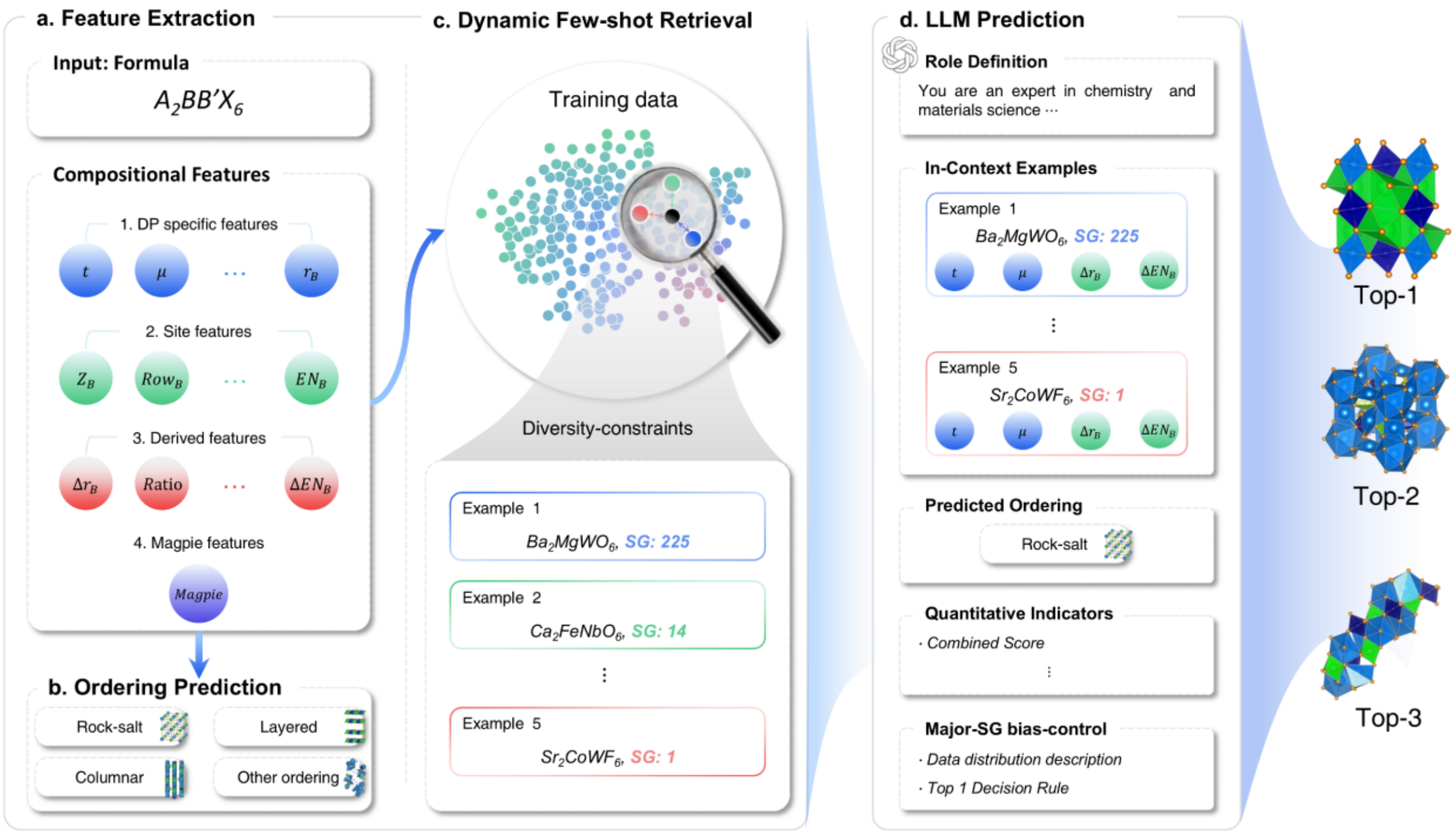

**Figure 1.** Schematic overview of DyRIS for SG prediction in DPs. (a) A query DP composition is transformed into composition-based descriptors, including DP-specific, site, *B*-site-derived, and Magpie features. (b) These descriptors are also used to predict *B/B′* cation ordering probabilities as structural prior information. (c) Diversity-enhanced dynamic few-shot retrieval selects compositionally relevant in-context examples in the weighted embedding space while maintaining SG-label diversity. (d) The LLM agent integrates the retrieved examples, ordering prior, quantitative indicators and major-SG bias-control to perform rule-guided inference and generate ranked Top-3 SG candidates.

**Table 1.** Configuration of the LLM-agent component in DyRIS. The table lists the model backbone, API identifier, access mode, agent framework, and temperature setting used in this study.

| Item | Setting |
|---|---|
| LLM backbone | gpt-5.4-mini-2026-03-17 |
| Access mode | OpenAI API |
| Agent framework | Microsoft AutoGen |
| Temperature | 1.0 |
| Setting mode | API/framework default; not explicitly specified |

### 2.1 Data preprocessing

The energy above the convex hull, $E_{hull}$, is widely used as an indicator of thermodynamic stability. A material with $E_{hull} \leq 0.1\ eV/atom$ is commonly regarded as a metastable candidate.[36] To construct the dataset, we retained only DP candidates with $E_{hull} \leq 0.1\ eV/atom$. For compositions with multiple entries, the structure with the lowest $E_{\mathrm{hull}}$ was selected. After filtering this stability, SG classes containing three or fewer samples were excluded from both training and evaluation. Classes with very few samples provide unstable estimates of class centers, variances, and boundaries, which makes it difficult to construct a reliable decision boundary.[37] After filtering by $E_{hull}$ and class size, the final dataset contained 3,528 DPs across 19 SG classes. Two classes, SG 14 and SG 225, together account for 72% of the dataset, with 1,081 and 1,460 samples, respectively. We refer to these two classes, SG 14 and SG 225, as major SGs, and the remaining 17 classes as minor SGs. The SG-class distribution of the dataset is summarized in **Table 2**.

The 3,528 DPs were divided into training and test sets. The training set was used to implement DyRIS and train the baseline models, whereas the test set was used for performance evaluation. The training set was sampled using stratified random sampling according to a predefined training-data ratio, so that the SG distribution of the original dataset was preserved. The test set was constructed from the remaining samples by randomly selecting up to 30 samples per SG. This cap prevents overall accuracy from being dominated by major-SG samples. If fewer than 30 samples were available for a given SG after selecting the training set, all remaining samples from that SG were included in the test set.

**Table 2.** Space-group distribution of the final DP dataset after stability and class-size filtering. The table reports the number of collected DP entries assigned to each SG class. SG 14 and SG 225 contain the largest numbers of samples and are therefore defined as major SGs in this study, whereas all remaining SG classes are treated as minor SGs.

| **SG 1** | **SG 2** | **SG 7** | **SG 11** | **SG 12** | **SG 14** | **SG 26** | **SG 31** | **SG 34** | **SG 87** |
|---|---|---|---|---|---|---|---|---|---|
| 105 | 177 | 7 | 11 | 177 | 1081 | 14 | 30 | 6 | 32 |

| **SG 123** | **SG 139** | **SG 146** | **SG 148** | **SG 164** | **SG 166** | **SG 194** | **SG201** | **SG 225** |
|---|---|---|---|---|---|---|---|---|
| 12 | 45 | 158 | 21 | 89 | 71 | 4 | 28 | 1460 |

## 2.2 Feature extraction

Four categories of composition-based features were used in this study: 1) DP-specific features, 2) Site features, 3) $B$-site-derived features, and 4) Magpie features. All four feature groups were used to construct the embedding space for diversity-enhanced dynamic few-shot retrieval. They were also used to train the surrogate model for *B/B′* cation ordering prediction, which is incorporated into rule-guided SG inference. Detailed procedures and examples for feature construction are provided in the **Section S1, Supporting Information (SI).**

### 1) DP-specific features

The formation of DP structures is closely related to ionic size and charge balance. Oxidation states provide a basic criterion for charge neutrality, while Shannon ionic radii quantify ionic size by accounting for both oxidation state and coordination environment.[38,39] These quantities are therefore closely associated with structural formation and distortion in DPs. In addition, the tolerance factor $t$ and octahedral factor $\mu$, derived from ionic radii and oxidation states, are widely used descriptors of perovskite stability, octahedral tilting, and symmetry lowering.[40] Based on these considerations, we selected the tolerance factor $t$, octahedral factor $\mu$, Shannon ionic radii, and oxidation states as DP-specific features.

### 2) Site features

Even when two DPs have similar composition-level statistics, different elemental arrangements at the $A$, $B$, $B'$, and $X$ sites can lead to different SGs. It is therefore useful to represent site-specific chemical information rather than relying only on composition-level statistics. In addition, the $A$-site cation can affect tilting modes and phase stability, which are directly related to SG symmetry.[41] Accordingly, the site feature group was constructed using the atomic number, period, group, and electronegativity of each crystallographic site.

**3) *B*-site-derived features**

*B/B′* cation ordering is one of the key structural characteristics that distinguishes DPs from simple perovskites.[17] Ordering tendencies and local structural distortions are influenced by differences in oxidation state, ionic radius, and electronegativity between the *B* and *B′* cations.[42] Because *B/B′* ordering constrains possible crystal symmetries and SG candidates, these relative differences are important for SG prediction.[17] To capture this information, we selected the oxidation-state difference, ionic-radius difference, and electronegativity difference between the *B* and *B′* cations as *B*-site-derived features.

**4) Magpie features**

The overall chemical characteristics of a composition can also influence structural information such as *B/B′* ordering, crystal system, and SG.[43,44] To represent the broader chemical environment, we used Magpie features, where Magpie denotes the Materials Agnostic Platform for Informatics and Exploration.[45] Magpie features quantify composition-level chemical trends by computing statistical descriptors, such as means, ranges, and variances, from elemental properties. These composition-based descriptors have been widely used in inorganic materials prediction tasks and were incorporated in this study to supplement the chemical information that may not be fully captured by the DP-specific and site-specific descriptors.[46]

### 2.3 Diversity-enhanced dynamic few-shot prompting

We propose diversity-enhanced dynamic few-shot prompting to address SGs imbalance in DP datasets. The approach operates in an embedding space, where each composition is represented as a numerical feature vector. Rather than relying on a fixed set of in-context examples—training samples inserted into the LLM input as references—the method dynamically selects examples for each query. Specifically, it retrieves query-specific examples based on the position of the query composition in the embedding space and incorporates them into the prompt.

A conventional k-nearest-neighbor (kNN)-based retrieval strategy selects the nearest training samples to the query composition as in-context examples.[47] However, in an imbalanced dataset, this approach can become biased toward nearby samples from major SGs.[48] To reduce this bias, DyRIS restricts the retrieved example set so that at most one example is selected from each SG. This diversity constraint prevents the prompt from being dominated by major SGs[49,50] and increases the probability that minor SG candidates are included in the retrieved set. We refer to the overall strategy as diversity-enhanced dynamic few-shot prompting and to the example-selection step as "Diversity-enhanced dynamic few-shot retrieval". For brevity, the example-selection step is referred to as "Diversity-enhanced retrieval" in the following sections.

We also introduced feature-block weighting, in which weights were determined from the training data, to construct a more informative embedding space. The four feature blocks—DP-specific features, Site features, *B*-site-derived features, and Magpie features—capture different levels of structural and chemical information. A simple concatenation of these feature blocks may not reflect their relative importance for SG prediction.[51] Therefore, we optimized the weights assigned to each feature block using Overall Top-1 macro-F1 score as the performance criterion within the training data.[52] This weighted embedding was used for the retrieval step in DyRIS. Details of the weight-search procedure and the resulting performance improvement are provided in **Section S2, SI.**

### 2.4 Rule-guided LLM-based SG inference

Retrieval obtained from Diversity-enhanced retrieval alone is insufficient to determine the final Top-3 SG candidates and their ranking. Comparing the candidates requires several quantitative indicators, whose definitions are provided in Section 2.4.2. Because these indicators have different meanings and numerical scales, it is difficult to process them uniformly using a single fixed threshold or a fixed set of weights. More importantly, the relative importance of each indicator depends on the query; therefore, candidate selection and ranking cannot be adequately handled by a fixed rule applied identically to all queries.

DyRIS therefore uses the LLM not as a simple predictor but as an inference agent that integrates multiple sources of evidence. The rule-guided LLM determines the final SG candidates and their ranking by jointly considering the information provided in the prompt and query, including *B/B′* cation ordering and candidate-specific quantitative indicators. In the following subsections, we describe the information provided in the prompt and query to support SG inference, including *B/B′* cation ordering and quantitative indicators in Sections 2.4.1 and 2.4.2, respectively, followed by the rule-guided inference procedure in Section 2.4.3.

#### 2.4.1 *B/B′* cation ordering

*B/B′* cation ordering provides important prior structural information for SG prediction in DPs.[43,53] *B/B′* ordering in double perovskites is commonly classified into rock-salt, layered, and columnar arrangements, among which rock-salt ordering is the most frequently observed.[43] In this study, non-rock-salt cases, including layered, columnar, and other ordering that cannot be assigned to any of the three ordered patterns, are collectively referred to as rare ordering.

DyRIS uses a PyCaret-based surrogate model to predict the probabilities of *B/B′* ordering types for each query composition.[54] The construction and performance of this ordering surrogate model are described in **Section S3, SI**. These predicted probabilities are passed to the LLM as part of the query. The system prompt separately provides a mapping table linking each ordering type to its compatible SG set. Using this mapping, the query also flags which of the retrieved Top-5 SG candidates are compatible with each ordering type.

Because rare ordering appears infrequently in the training data, its predicted probability can be smaller than that of rock-salt ordering even when rare-ordering-compatible SGs are plausible. Therefore, DyRIS does not use the most probable ordering type as a deterministic label. Instead, the ordering probabilities and compatible SG candidates are provided to the LLM as soft structural prior information. The rationale for employing the PyCaret-based prediction model and a more detailed analysis of the discriminative power of ordering information are provided in **Section S3, SI.**

### 2.4.2 Quantitative indicators

Quantitative indicators were introduced to provide explicit numerical criteria that guide the LLM agent during the inference process, enabling more reliable selection of the final Top-3 SG candidates and their ranking by reducing ambiguity and supporting consistent decision-making.

These indicators fall into two complementary groups. The first is a density-based indicator, the Combined score, which measures how frequently a candidate SG appears among the training samples nearest to the query composition and thereby prioritizes candidates that recur in its immediate neighborhood. The second group is distribution-based and is derived from the standardized deviation between the query composition and the feature distribution of each candidate SG. We define the Global fit as a distance-like measure that aggregates these deviations across six representative features to support the final Top-1 ranking, whereas the feature-wise consistency metrics quantify the average agreement, the worst-case mismatch, and the number of individual features for which each candidate SG shows the best agreement with the query.

The Combined score quantifies how densely each candidate SG appears in the consecutive nearest-neighbor region around the query composition. It was introduced to prioritize SG candidates that are repeatedly observed near the query composition:

$$CS(s) = log(1 + n_{local}(s)) \cdot \frac{n_{local}(s)}{N_s + 1} \qquad (1)$$

Here, *CS* denotes the Combined score, and *s* denotes one of the SG candidates obtained from the five retrieved in-context examples. $n_{\mathrm{local}}(s)$ is defined as the number of consecutive nearest-neighbor training samples that share candidate SG *s* label. To obtain it, all training samples are first sorted by their distance to the query composition, starting from the first occurrence of *s* in this sorted list. We count the consecutive samples assigned to *s* until a different SG label appears. $N_s$ denotes the total number of training samples belonging to SG *s*, and the term $N_s+1$ is used to reduce the effect of globally frequent SGs. The *CS* follows the instance-based learning principle that classes with higher local density around a query are more reliable candidates.[55,56] At the same time, normalization by $N_s+1$ prevents candidate SGs from being overestimated simply because they are globally frequent in the training data.

To define the remaining quantitative indicators, we first compute the standardized deviation between the query composition and the feature distribution of each candidate SG:

$$z(s,f) = \frac{x_f - \mu(s,f)}{\sigma(s,f)} \qquad (2)$$

Here, $x_f$ is the value of feature $f$ for the query composition, and $\mu(s,f)$ and $\sigma(s,f)$ are the mean and standard deviation, respectively, of feature $f$ among training samples assigned to SG *s*. A smaller $|z(s,f)|$ indicates that the query composition is closer to the feature distribution of candidate SG *s*. This follows the same principle as standardized-distance- and Mahalanobis-distance-based group assignment in chemometrics.[57]

The Global fit evaluates how well a candidate SG matches the query composition across multiple embedding features. It is defined as a distance-like quantity based on six standardized feature values:

$$G_{fit}(s) = \sqrt{z(s,tf)^2 + z(s,of)^2 + z(s,r_A)^2 + z(s,r_{B'}/r_B)^2 + z(s,\Delta r_B)^2 + z(s,\Delta EN_B)^2} \qquad (3)$$

Here, $z(s,tf), z(s,of), z(s,r_A), z(s,r_{B'}/r_B), z(s,\Delta r_B)$ and $z(s,\Delta EN_B)$ denote the standardized deviations of the tolerance factor, octahedral factor, A-site ionic radius, *B/B′* ionic radius ratio, B-site radius difference, and B-site electronegativity difference, respectively, between the query composition and the training distribution of candidate SG $s$. A lower Global fit indicates better agreement between the query composition and the feature distribution of the candidate SG. This indicator compresses multiple feature-level deviations into a single overall distributional-fit measure and is used to support the final Top-1 ranking among the selected Top-3 SG candidates.

Although Global fit provides an overall distance-based measure by integrating standardized deviations across multiple embedding features, it does not explicitly indicate how each individual feature contributes to the final ranking. Therefore, feature-wise consistency metrics were introduced to examine the agreement between the query composition and each candidate SG at the individual feature level:

$$z_{abs_mean}(s) = \frac{1}{|F|}\sum_{f \in F} |z(s,f)| \qquad (4)$$

$$z_{max_abs}(s) = \max_{f \in F} |z(s,f)| \qquad (5)$$

$$z_{best_count}(s) = \sum_{f \in F} 1\{|z(s,f)| = \min_{s\prime \in C} |z(s',f)|\} \qquad (6)$$

Here, $F$ denotes the feature set used for SG comparison and $|F|$ is the number of features in this set. $C$ is the retrieved candidate SG set, and $z(s,f)$ is the standardized deviation of candidate SG $s$ for feature $f$. $\mathbf{1}\{\cdot\}$denotes the indicator function, which equals 1 if the condition is true and 0 otherwise.

The metric $z_{\text{abs_mean}}(s)$ represents the average magnitude of feature-wise deviation, thereby quantifying the overall feature-level agreement between the query and a candidate SG. A smaller value indicates that the query composition is, on average, closer to the feature distribution of the candidate SG.

The metric $z_{\text{max_}abs}(s)$ captures the largest feature-wise deviation and is therefore used to identify potential feature-specific mismatch. A smaller value implies that the candidate SG does not exhibit a pronounced discrepancy in any individual feature.

In contrast, $z_{\text{best_count}}(s)$ measures the number of features for which a given candidate SG shows the smallest absolute standardized deviation among the retrieved candidates. This metric reflects the relative feature-level dominance of a candidate within the candidate set. $z_{\text{best_count}}(s)$ evaluates how frequently the candidate provides the closest feature-wise match compared with the alternatives.

They allow the final Top-1 selection among the Top-3 SG candidates to account not only for overall distributional agreement, but also for average consistency, worst-feature mismatch, and relative feature-wise superiority.

### 2.4.3 Core inference rules for rule-guided inference

To transform the retrieved Top-5 SG candidates into the final Top-3 SG prediction, DyRIS follows a rule-guided inference procedure that integrates retrieval evidence, ordering compatibility, Global fit, and feature-wise consistency. The detailed step-by-step procedure, full prompt, and query examples are provided in **Section S4, SI**.

The rule-guided inference procedure is based on five principles. First, candidate SGs with high Combined scores are preferentially retained in the initial Top-3 candidate set because they are repeatedly supported by retrieved examples similar to the query composition. Second, candidate replacement is performed only when ordering compatibility, Global fit, feature-wise consistency, and retrieval-distance ranking provide consistent evidence. Third, rare-ordering-compatible SGs are considered for inclusion in the Top-3 candidate set when they are supported by both ordering evidence and feature-level agreement. Fourth, major-SG bias control is applied to mitigate the over-selection of SG 14 and SG 225, which are overrepresented in the training data. Specifically, these SGs are assigned to the Top-1 position only when the Combined score, retrieval-distance ranking, Global fit, and feature-wise z-score consistency all provide clear support. Finally, after the final candidate set is determined, the final ranking is assigned by jointly considering the Combined score, retrieval-distance ranking, Global fit, and feature-wise consistency.

Thus, DyRIS does not rely on a single score or a fixed weighted sum. Instead, it produces the final Top-3 SG prediction by structurally integrating query-specific evidence within the rule-guided inference procedure.

## 3. Results and discussion

We performed four experiments to evaluate the performance and characteristics of DyRIS. All experiments evaluated both the full test set and the minor-SG subset, which excludes SG 14 and SG 225. All F1 scores are reported as macro-averaged F1 scores. The metrics evaluated on the full test set are denoted as Overall Top-1 accuracy, Overall Top-1 macro-F1 score, and Overall Top-3 accuracy. The corresponding metrics for the minor-SG subset are denoted as Minor-SG Top-1 accuracy, Minor-SG Top-1 macro-F1 score, and Minor-SG Top-3 accuracy. This metric design allows the framework to be evaluated not only on overall prediction performance but also on minor-SG performance, which is strongly affected by class imbalance.

### 3.1 Performance of DyRIS

#### 3.1.1 Model comparison

To evaluate the prediction performance of DyRIS under a fixed training-data ratio, we set the ratio to 0.5. DyRIS was compared with the best-performing PyCaret-based model, CrabNet, and CRYSPNet under the same train/test split. CrabNet and CRYSPNet were selected as representative data-driven models for composition-based and descriptor-based SG prediction.[23,58] The data split was repeated five times with different random seeds, and all performance values were reported as mean ± standard deviation.

As shown in **Figure 2**, DyRIS showed the most balanced performance across the evaluated metrics. For the overall SG set, DyRIS achieved Top-1 accuracy comparable to CrabNet, which was the strongest baseline for this metric, and outperformed the other baselines in Overall Top-1 macro-F1 score (**Figure 2(a) and (b)**). This indicates that DyRIS maintains competitive Top-1 accuracy while reducing the class-imbalance effect reflected in macro-F1. In Overall Top-3 accuracy, the PyCaret-based model showed the highest performance, but DyRIS remained competitive and outperformed CrabNet and CRYSPNet (**Figure 2(c)**).

The advantage of DyRIS became clearer in the minor-SG subset. As shown in **Figure 2(d–f)**, DyRIS achieved the best performance in Minor-SG Top-1 accuracy, Minor-SG Top-1 macro-F1 score, and Minor-SG Top-3 accuracy. This result indicates that DyRIS does not improve performance simply by favoring majority SGs, but provides more reliable predictions for underrepresented SG classes. In contrast, CrabNet was mainly strong in Overall Top-1 accuracy, and the PyCaret-based model was mainly strong in Overall Top-3 accuracy, whereas their minor-SG performance was lower than that of DyRIS.

Overall, these results show that DyRIS is distinguished from the baseline models by its ability to balance overall prediction performance with improved minor-SG prediction. This balanced behavior is particularly important for DP SG prediction, where the training distribution is strongly imbalanced.

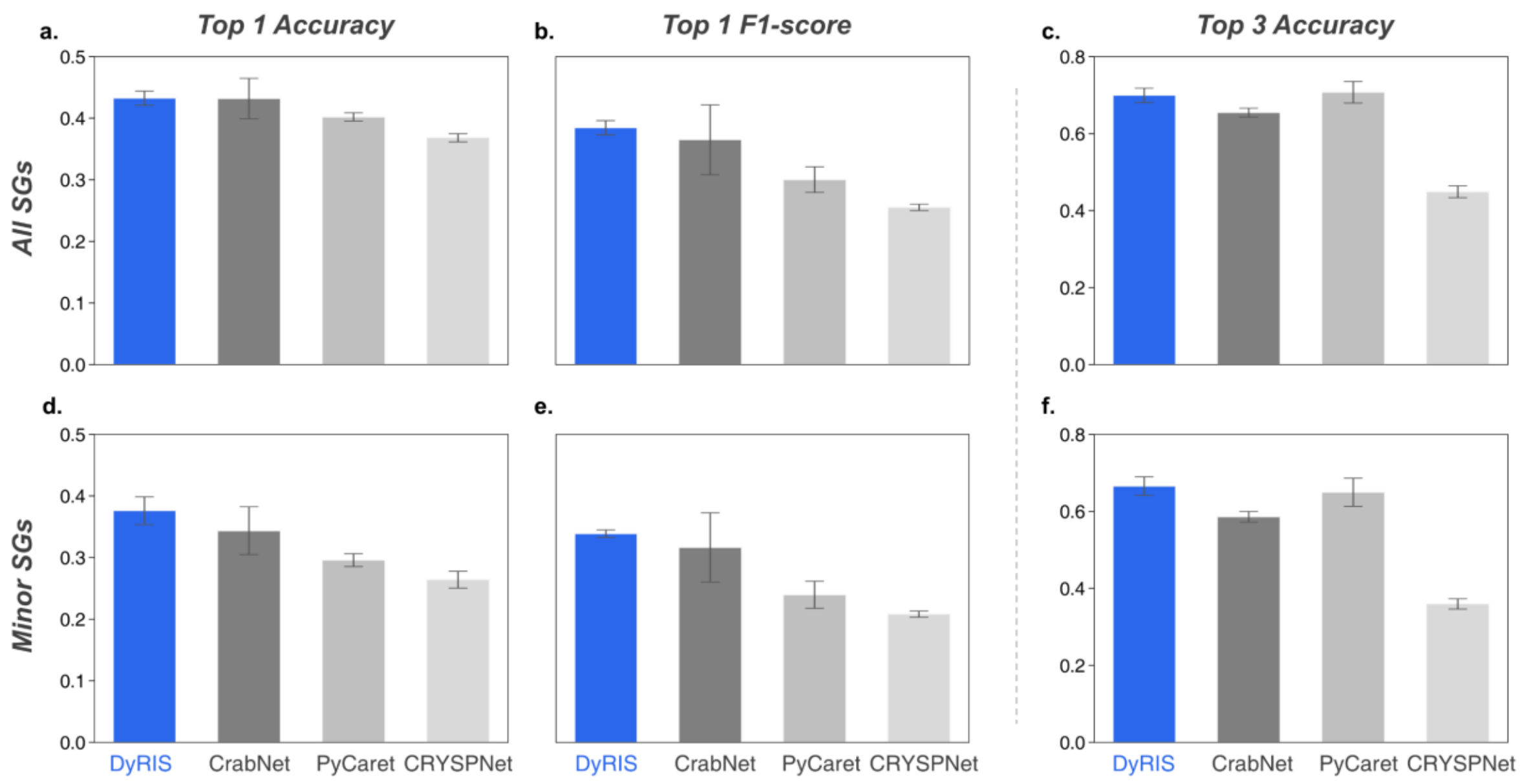

**Figure 2.** Comparison of DyRIS with baseline models at a training-data ratio of 0.5. (a–c) Overall Top-1 accuracy, Top-1 macro-F1 score, and Top-3 accuracy evaluated on all SG classes. (d–f) Corresponding metrics evaluated on the minor-SG subset, excluding SG 14 and SG 225. Error bars represent the standard deviation over five random train/test splits. DyRIS maintains competitive overall accuracy while achieving improved performance in Overall Top-1 macro-F1 score and all minor-SG metrics.

### 3.1.2 Data sensitivity across training-data ratios

To investigate how the performance of DyRIS changes with the amount of training data, we varied the training-data ratio from 0.3, 0.5, and 0.8. DyRIS was compared with two baseline models selected based on the fixed-ratio model comparison described in Section 3.1.1. For each training-data ratio, the data split was repeated five times with different random seeds. CrabNet and the PyCaret-based model were selected as the main baselines because they showed strong Top-1 and Top-3 performance, respectively, in Section 3.1.1.

**Table 3.** Performance comparison of (a) DyRIS, (b) the PyCaret-based model, and (c) CrabNet across different training-data ratios. The table reports Overall Top-1 accuracy, Overall Top-1 macro-F1 score, Overall Top-3 accuracy, Minor-SG Top-1 accuracy, Minor-SG Top-1 macro-F1 score, and Minor-SG Top-3 accuracy at training-data ratios of 0.3, 0.5, and 0.8. Values represent the mean performance over five random train/test splits.

(a) DyRIS

| | **Overall SGs** | | | **Minor SGs** | | |
|---|---|---|---|---|---|---|
| Ratio | Top1 acc | Top1 F1 | Top3 acc | Top1 acc | Top1 F1 | Top3 acc |
| 0.3 | 0.3598 | 0.3074 | 0.6264 | 0.3132 | 0.2706 | 0.5834 |
| 0.5 | 0.4266 | 0.3846 | 0.6994 | 0.3762 | 0.3388 | 0.6658 |
| 0.8 | 0.4154 | 0.3632 | 0.7334 | 0.3534 | 0.3266 | 0.6814 |

(b) PyCaret-based model

| | **Overall SGs** | | | **Minor SGs** | | |
|---|---|---|---|---|---|---|
| Ratio | Top1 acc | Top1 F1 | Top3 acc | Top1 acc | Top1 F1 | Top3 acc |
| 0.3 | 0.343 | 0.2326 | 0.6434 | 0.2494 | 0.1956 | 0.5998 |
| 0.5 | 0.402 | 0.3002 | 0.7074 | 0.2958 | 0.2396 | 0.6494 |
| 0.8 | 0.4588 | 0.3046 | 0.7704 | 0.3056 | 0.2380 | 0.7022 |

(c) CrabNet

| | **Overall SGs** | | | **Minor SGs** | | |
|---|---|---|---|---|---|---|
| Ratio | Top1 acc | Top1 F1 | Top3 acc | Top1 acc | Top1 F1 | Top3 acc |
| 0.3 | 0.363 | 0.2806 | 0.5886 | 0.2756 | 0.2358 | 0.5188 |
| 0.5 | 0.4318 | 0.365 | 0.6546 | 0.3436 | 0.3164 | 0.5860 |
| 0.8 | 0.4666 | 0.361 | 0.7028 | 0.3424 | 0.2936 | 0.5940 |

As shown in **Table 3** and **Figure 3(a)**, DyRIS achieved Overall Top-1 accuracy comparable to CrabNet at training-data ratios of 0.3 and 0.5, but its advantage weakened at a training-data ratio of 0.8. At this highest ratio, DyRIS achieved an Overall Top-1 accuracy of 0.4154, which was lower than that of CrabNet (0.4666) and the PyCaret-based model (0.4588). In contrast, **Figure 3(b)** shows that DyRIS achieved the highest Overall Top-1 macro-F1 score at all training-data ratios, indicating that it maintained the most balanced performance across SG classes under class imbalance.

The advantage of DyRIS was more consistent for minor-SG prediction. As shown in **Figures 3(d) and 3(e)**, DyRIS achieved the highest Minor-SG Top-1 accuracy and Minor-SG Top-1 macro-F1 score at all training-data ratios. These results indicate that DyRIS improves prediction performance not simply by favoring major SGs, but by providing more reliable predictions for minor SGs classes.

For Top-3 prediction, as shown in **Table 3**, the PyCaret-based model showed the strongest overall performance at all training-data ratios, while DyRIS consistently ranked second and outperformed CrabNet. For Minor-SG Top-3 accuracy, DyRIS outperformed the PyCaret-based model at a training-data ratio of 0.5 and remained competitive at the other ratios. This indicates that DyRIS remains effective at including the correct SG within the Top-3 candidates, especially for minor-SG samples.

Overall, DyRIS consistently achieved strong class-balanced and minor-SG performance across training-data ratios, while maintaining competitive overall accuracy. However, increasing the training-data ratio did not uniformly improve all DyRIS metrics. In particular, several Top-1 metrics decreased when the training-data ratio increased from 0.5 to 0.8. This limitation is analyzed in detail in Section 3.3.1. The complementary behavior of DyRIS and the PyCaret-based model also motivates the hybrid strategy discussed in Section 3.3.4.

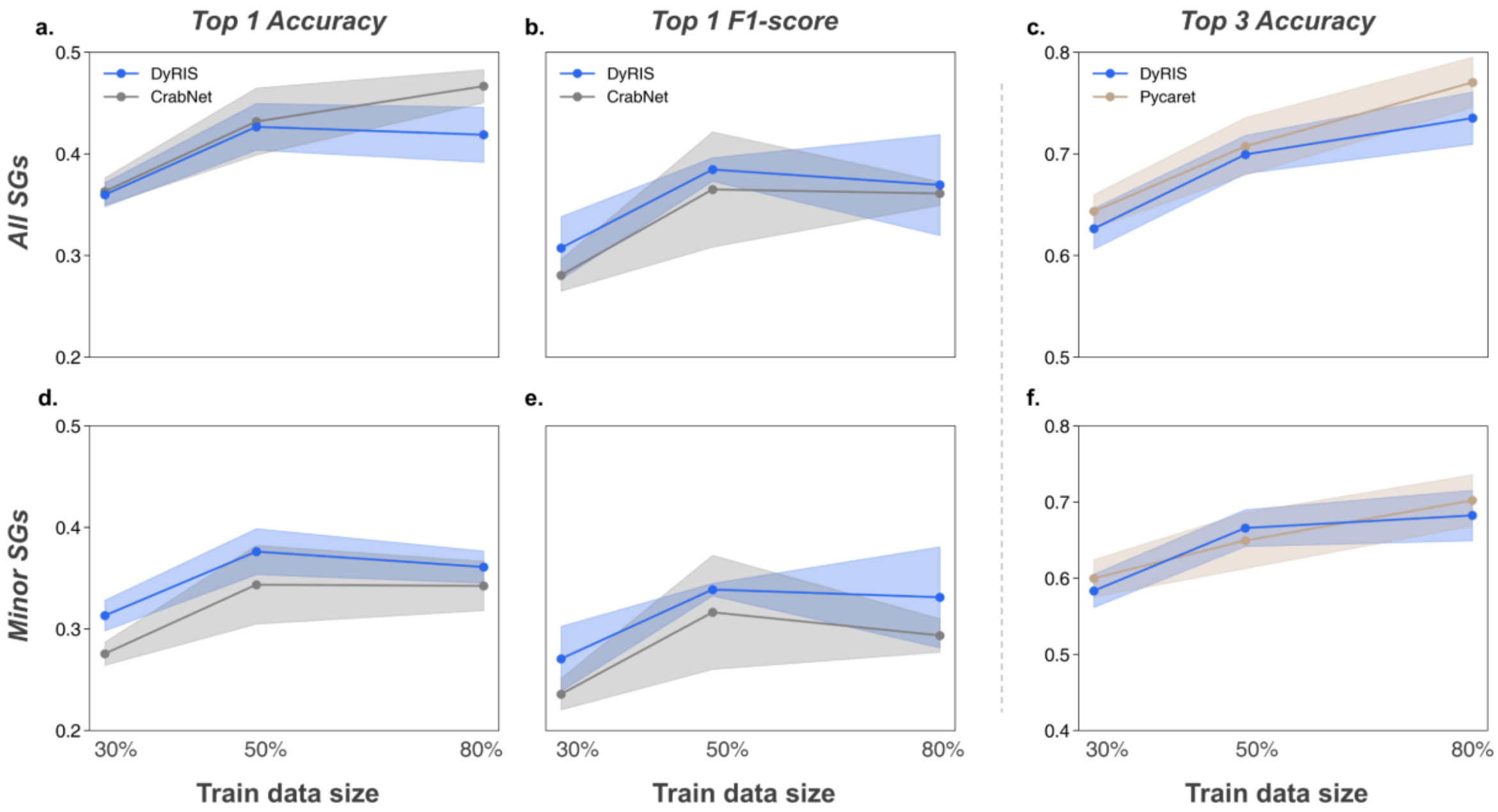

**Figure 3.** Effect of training-data ratio on the performance of DyRIS. (a–c) Overall Top-1 accuracy, Top-1 macro-F1 score, and Top-3 accuracy at training-data ratios of 30%, 50%, and 80%. (d–f) Corresponding metrics evaluated on the minor-SG subset. For Top-1 metrics, DyRIS is compared with CrabNet, whereas for Top-3 metrics, DyRIS is compared with the PyCaret-based model. Shaded regions represent the standard deviation over five random train/test splits.

### 3.2 Analysis of the performance gains in DyRIS

#### 3.2.1 Ablation study

An ablation study was conducted to analyze the contribution of each component in DyRIS. Starting from the full DyRIS framework, we removed in-context examples, ordering information, quantitative indicators, and the major-SG bias-control, respectively. We then evaluated the resulting performance under the same train/test split. All ablation experiments were performed at a training-data ratio of 0.5 and repeated over five random splits.

The largest performance degradation occurred when the retrieval-based in-context examples were removed. As shown in **Figure 4(a)**, overall Top-3 accuracy decreased from 0.6994 to 0.5424, indicating that retrieval-based in-context examples are a key component for narrowing the possible SG candidate space to plausible candidates. The removal of in-context examples also strongly affected Top-1 prediction, as Overall Top-1 accuracy decreased from 0.4266 to 0.2214. These results indicate that failure to construct an appropriate Top-3 candidate set propagates directly to the final Top-1 prediction. This effect was also pronounced for minor SGs, as shown in **Figure 4(b)**.

We also evaluated the retrieval-only setting, where the retrieved SG candidates were directly used as predictions without LLM-based candidate refinement and ranking. The retrieval-only performance was lower than that of full DyRIS, indicating that retrieval alone is insufficient for final SG ranking. This result suggests that, although retrieval-based examples are essential for constructing a plausible candidate set, additional evidence is required to discriminate among the retrieved candidates. Detailed retrieval-only results are provided in **Section S5, SI.**

Removing the quantitative indicators also led to consistent decreases across all metrics, with larger degradation observed in the minor-SG subset. For example, Minor-SG Top-1 accuracy decreased by approximately 5.5 percentage points. This behavior is related to the limited number of training samples available for minor SGs. When only a small number of samples are available, predictions based mainly on retrieval evidence can have high variance and can be easily affected by neighboring major SGs or other minor SGs in the embedding space. DyRIS reduces this instability by providing multiple forms of quantitative evidence. These indicators are therefore inferred to contribute to more stable candidate selection and ranking for minor SGs, which is difficult to achieve using retrieved examples alone.

Removing the major-SG bias-control also led to performance degradation. In this ablation, we removed the prompt information that explicitly identified SG 14 and SG 225 as majority classes and the instruction that these SGs should be selected as Top-1 only when strongly supported by multiple indicators. As shown in **Figure 4(a),** Overall Top-1 accuracy and Overall Top-1 macro-F1 score decreased by approximately 1.0 and 3.0 percentage points, respectively. A similar decrease was also observed for the minor-SG subset, as shown in **Figure 4(b).** This implies that the major-SG bias-control is likely to support the final ranking step by preventing over-ranking of majority SGs and preserving minor-SG candidates.

Removing ordering information at a training-data ratio of 0.5 produced only small changes, and these changes were not consistently positive or negative across metrics. This indicates that, under this data condition, the contribution of ordering information was not apparent. To further examine whether ordering information provides a consistent benefit, we performed an additional ablation analysis at a training-data ratio of 0.8. The effect of ordering information became more evident in this high-data condition. Removing ordering information decreased most metrics, with Minor-SG Top-1 accuracy decreasing by approximately 4.2 percentage points, as shown in **Figure 4(c).** This trend can be interpreted as indicating that, as the training pool becomes larger, ordering compatibility more frequently helps distinguish boundary candidates within the retrieved Top-5 candidate set. Therefore, ordering information acts as a structural prior that adjusts the final candidate set toward crystallographically plausible SGs. Additional analyses of ordering ablation, selectivity, and actionability within the retrieved Top-5 candidate set are provided in **Section S6, SI.**

Overall, the performance of DyRIS originates from the integration of these four components. Retrieval-based in-context examples play the primary role of narrowing the possible SG candidate space. Quantitative indicators improve candidate selection and ranking by reflecting both the local density of retrieved SG candidates around the query composition and their distributional agreement with the query features. The major-SG bias-control helps reduce over-ranking of majority SGs during final ranking, thereby supporting more balanced predictions for minor SGs. Ordering information contributes as a structural prior that guides the prediction toward crystallographically plausible candidates, with a stronger effect in the high-data regime. Thus, the performance improvement of DyRIS does not arise from a single rule or a single feature, but from the integrated inference of retrieval evidence, quantitative evidence, major-SG bias control, and structural prior information.

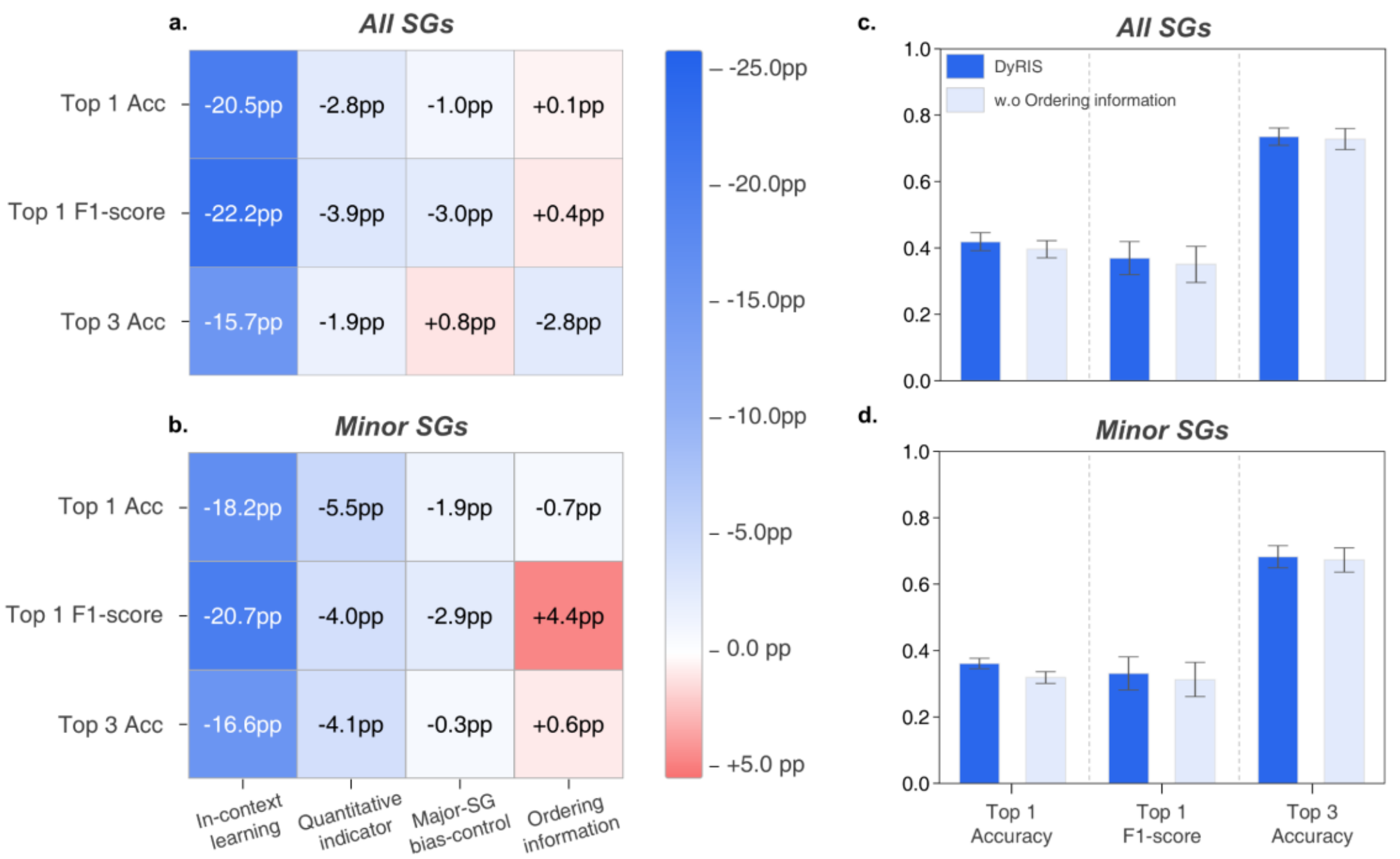

**Figure 4.** Ablation analysis of DyRIS. (a, b) Performance changes caused by removing each component of DyRIS at a training-data ratio of 0.5 for (a) all SGs and (b) the minor-SG subset. The heatmaps show the percentage-point change relative to the full DyRIS model after removing retrieval-based in-context examples, quantitative indicators, the major-SG bias-control, or ordering information. (c, d) Comparison between full DyRIS and DyRIS without ordering information at a training-data ratio of 0.8 for (c) all SGs and (d) the minor-SG subset. The bar plots report Top-1 accuracy, Top-1 F1-score, and Top-3 accuracy, with error bars indicating the standard deviation over five random train/test splits.

### 3.2.2 Classifier-based replacement of rule-guided inference

To examine whether the final rule-guided inference step of DyRIS can be replaced by conventional learned re-ranking models, we compared DyRIS with classifier- and ranker-based alternatives while keeping the retrieval stage and candidate evidence construction unchanged. In this experiment, the diversity-enhanced retrieval stage of DyRIS was retained, and only the final Top-3 candidate selection and ranking step was replaced with ML based re-ranking models.

For each query, candidate-level evidence was constructed from the retrieved Top-5 SG candidates, including retrieval rank, Combined score, Global fit, feature-wise z-score consistency indicators, and ordering-compatibility features. To prevent information leakage, training evidence was generated using a leave-one-out protocol within each training split, where the query itself was excluded from both the retrieval pool and the computation of SG-wise normalization statistics. For test queries, candidate-level evidence was computed using only the corresponding training split. Using this evidence, we trained representative classifier-based models and a learning-to-rank model, including logistic regression, random forest, gradient boosting, LightGBM, XGBoost, and LambdaMART. The same train/test splits used in Section 3.1.2 were adopted for training-data ratios of 0.3, 0.5, and 0.8.

As shown in **Table 4**, DyRIS outperformed the best ML-based re-ranking model across all evaluated metrics and training-data ratios. The advantage was most pronounced in the minor-SG subset, indicating that the rule-guided inference step contributes more effectively to class-imbalanced prediction than supervised re-ranking using the same candidate-level evidence.

Although the performance gap was smaller for Top-3 accuracy, DyRIS still achieved higher Overall and Minor-SG Top-3 accuracy at all training-data ratios. This suggests that ML-based re-ranking can partially identify plausible candidate sets, but it is less effective at assigning the correct SG to the first rank. Therefore, the final DyRIS inference step is not simply replacing a learnable scoring function; rather, it integrates retrieval evidence, quantitative indicators, ordering compatibility, and major-SG bias control in a way that improves final candidate ranking, especially for minor SGs.

The best ML model also varied depending on the metric and training-data ratio. For example, at a training-data ratio of 0.3, logistic regression was the best ML model for Top-1 metrics, whereas the LightGBM ranker performed best for Top-3 metrics. At a training-data ratio of 0.5, XGBoost was best for Top-1 accuracy, random forest was best for Top-3 accuracy, and histogram gradient boosting was best for macro-F1 score. At a training-data ratio of 0.8, XGBoost was strongest for Top-1 metrics, whereas histogram gradient boosting was strongest for Top-3 metrics. This variation indicates that classifier- and ranker-based re-ranking is sensitive to the metric, training-data ratio, and model choice. In contrast, DyRIS used the same rule-guided inference strategy across all ratios while maintaining stronger Top-1 and minor-SG performance.

These results suggest that the final inference step of DyRIS is not easily reducible to a fixed threshold rule or a learned classifier over quantitative indicators. Although ML models can use candidate-level evidence to predict whether each candidate corresponds to the correct SG, they can still form decision boundaries biased toward major SGs under severe class imbalance. They may also fail to capture subtle differences among candidate SGs for minor SGs, where training samples are limited. In contrast, rule-guided inference in DyRIS jointly considers retrieval-distance ranking, the local density of candidate SGs around the query composition, distributional agreement between the query composition and each candidate SG, and ordering-based structural prior information. Therefore, the performance gain of DyRIS appears to be associated with the inference process that integrates candidate-level quantitative evidence with crystallographic prior information.

**Table 4.** Comparison between DyRIS and the best ML model obtained from classifier- and ranker-based replacement of the final inference step. The other components of DyRIS, including diversity-enhanced dynamic few-shot retrieval, ordering information, and quantitative indicators, were kept unchanged, whereas the final rule-guided inference step was replaced by learned re-ranking models. Overall Top-1 accuracy, Overall Top-1 macro-F1 score, Overall Top-3 accuracy, Minor-SG Top-1 accuracy, Minor-SG Top-1 macro-F1 score, and Minor-SG Top-3 accuracy are reported at training-data ratios of 0.3, 0.5, and 0.8. Minor-SG metrics were evaluated on the minor-SG subset, excluding SG 14 and SG 225.

| | | **Overall SGs** | | | **Minor SGs** | | |
|---|---|---|---|---|---|---|---|
| Model | Ratio | Top1 acc | Top1 F1 | Top3 acc | Top1 acc | Top1 F1 | Top3 acc |
| DyRIS | 0.3 | 0.3598 | 0.3074 | 0.6264 | 0.3132 | 0.2706 | 0.5834 |
| Best learned | 0.3 | 0.2904 | 0.1916 | 0.6104 | 0.1785 | 0.1729 | 0.5446 |
| DyRIS | 0.5 | 0.4266 | 0.3846 | 0.6994 | 0.3762 | 0.3388 | 0.6658 |
| Best learned | 0.5 | 0.3445 | 0.2449 | 0.6734 | 0.2270 | 0.2237 | 0.6118 |
| DyRIS | 0.8 | 0.4154 | 0.3632 | 0.7334 | 0.3534 | 0.3266 | 0.6814 |
| Best learned | 0.8 | 0.3885 | 0.2706 | 0.7131 | 0.2207 | 0.2408 | 0.6304 |

## 3.3 Analysis of DyRIS limitations in the high-data regime

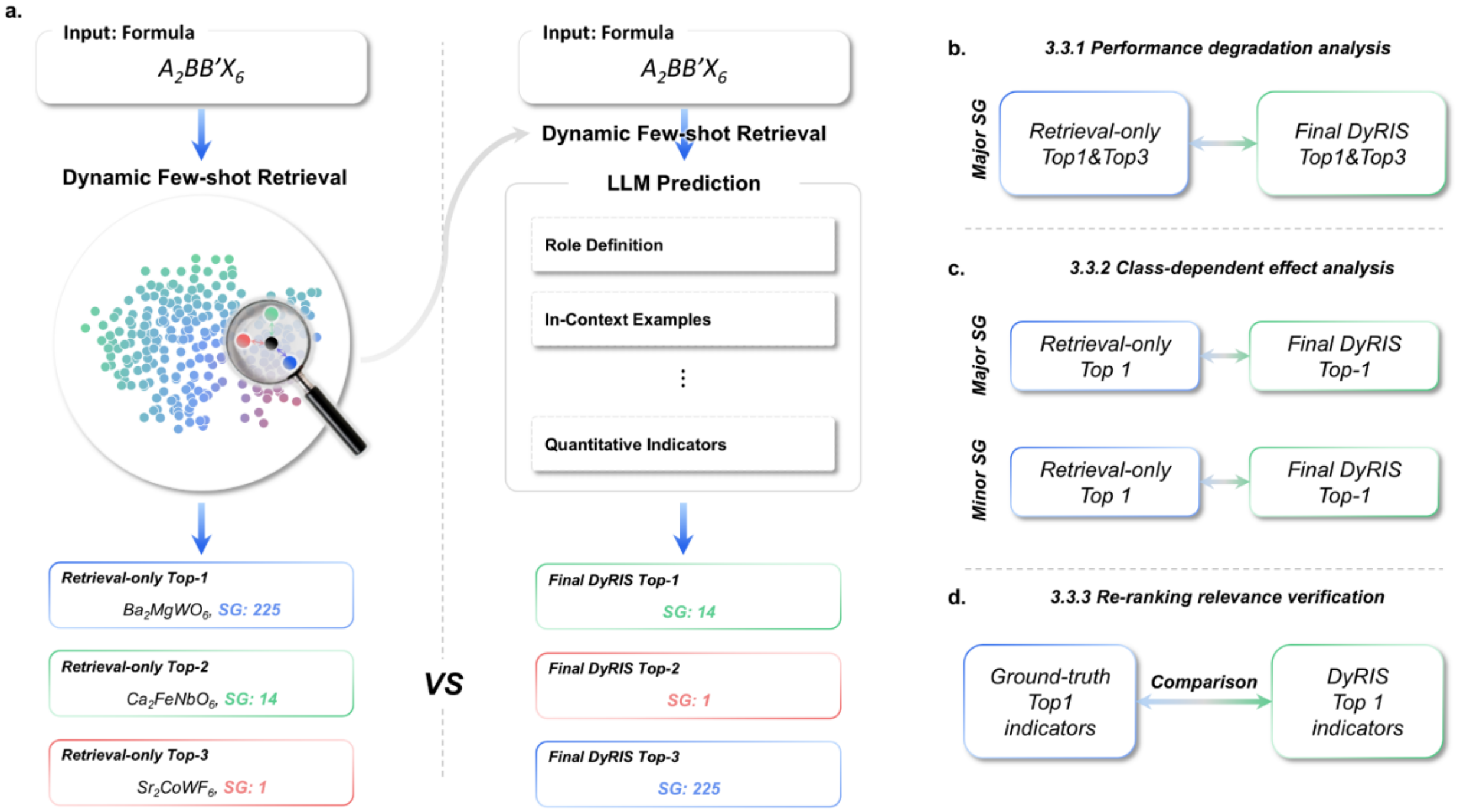

**Figure 5.** Schematic overview of the follow-up analyses used to interpret DyRIS prediction behavior. (a) Schematic illustration clarifying the retrieval-only prediction used in Section 3.3, in contrast to the final DyRIS prediction after LLM-based re-ranking. (b) Section 3.3.1 compares retrieval-only and final DyRIS Top-1/Top-3 performance to identify whether performance degradation arises from retrieval or final re-ranking. (c) Section 3.3.2 compares the re-ranking effect separately for major and minor SGs. (d) Section 3.3.3 verifies the relevance of the re-ranking process by comparing the quantitative indicators of the ground-truth Top-1 SG and the DyRIS-predicted Top-1 SG.

The preceding experiments showed that DyRIS achieves strong prediction performance for minor SGs, while also revealing a limitation in which several Top-1 metrics decrease in the high-data regime. Therefore, in Section 3.3, we performed a series of follow-up analyses to identify the source of this performance decrease and to examine how rule-guided inference operates (**Figure 5**). First, in Section 3.3.1, we separated the diversity-enhanced retrieval stage from the final ranking stage to determine whether the performance decrease was caused by a failure to include the correct SG in the candidate set or by the ranking step that selects the final Top-1 prediction from the retrieved candidates. In Section 3.3.2, we compared retrieval-only Top-1 prediction and final DyRIS Top-1 prediction separately for major SGs and minor SGs to identify which class type mainly contributed to the performance change during final ranking. In Section 3.3.3, we examined whether this class-dependent behavior was not arbitrary behavior of the LLM agent but instead reflected the quantitative indicators and ordering information provided in the prompt. Finally, in Section 3.3.4, based on the complementary strengths of DyRIS and the PyCaret-based model, we investigated the possibility of a hybrid approach that combines the two models.

### 3.3.1 Increasing training data does not always improve DyRIS performance

One limitation of DyRIS is that increasing the amount of training data does not consistently improve all performance metrics. As discussed in Section 3.1.2, several Top-1 metrics decreased when the training-data ratio increased from 0.5 to 0.8, including Overall Top-1 accuracy and Minor-SG Top-1 accuracy.

To identify where this degradation originates, we separated the retrieval stage from the final ranking stage. We first examined the Top-3 accuracy obtained using only diversity-enhanced retrieval, which measures whether the correct SG is included in the retrieved candidate set before rule-guided inference. When the training-data ratio increased from 0.5 to 0.8, retrieval-level Overall Top-3 accuracy increased from 0.6746 to 0.7213, and retrieval-level Minor-SG Top-3 accuracy increased from 0.6132 to 0.6402. The final Top-3 accuracy of DyRIS showed the same increasing trend, as shown in Section 3.1.2. Therefore, the decrease in Top-1 performance at a training-data ratio of 0.8 is unlikely to be caused by failure to include the correct SG within the candidate set. Instead, it is more likely associated with the final ranking step.

This interpretation is also consistent with the behavior of major SGs. For SG 14 and SG 225, the correct SGs were included in the candidate set more reliably as the training-data ratio increased, but their retention as the final Top-1 prediction decreased as shown in **Table 5**. This suggests that the final ranking step can sometimes demote the correct SG even when retrieval provides a plausible candidate set. The detailed mechanism behind this behavior, including the different effects of rule-guided inference on major SGs and minor SGs, is analyzed in Section 3.3.2.

The decrease in Overall Top-1 accuracy should also be interpreted together with the test-set composition. As summarized in **Table 6**, the total number of test samples decreases at a training-data ratio of 0.8, whereas SG 14 and SG 225 remain capped at 30 test samples each. As a result, these two major SGs occupy a larger fraction of the test set in the high-data regime. Under this evaluation condition, a decrease in Top-1 performance for SG 14 and SG 225 has a larger effect on Overall Top-1 accuracy. Taken together, these results indicate that the high-data limitation of DyRIS is mainly related to the final ranking step rather than the diversity-enhanced retrieval stage.

**Table 5.** Prediction accuracy for the major SGs, SG 14 and SG 225, at training-data ratios of 0.5 and 0.8. The table reports Top-1 accuracy and Top-3 accuracy for each major SG to examine how the final DyRIS ranking changes when the amount of training data increases. The analysis was performed using the same five data splits used in Section 3.1.2.

| | **Ratio=0.5** | | **Ratio=0.8** | |
|---|---|---|---|---|
| **SGs** | Top1 acc | Top3 acc | Top1 acc | Top3 acc |
| SG 14 | 0.4933 | 0.8200 | 0.4467 | 0.8733 |
| SG 225 | 0.8400 | 0.9000 | 0.7667 | 0.9133 |

**Table 6.** Test-sample counts for selected minor SGs across training-data ratios. The listed SG classes show the 5 largest decreases in test-sample count as the training-data ratio increases. These reductions explain why minor-SG metrics become more sensitive to individual predictions in the high-data regime.

| **Training ratio** | **SG 31** | **SG 87** | **SG 139** | **SG 166** | **SG 201** |
|---|---|---|---|---|---|
| 0.1 | 27 | 29 | 30 | 30 | 25 |
| 0.3 | 21 | 22 | 30 | 30 | 20 |
| 0.5 | 15 | 16 | 23 | 30 | 14 |
| 0.8 | 6 | 6 | 9 | 14 | 6 |

#### 3.3.2 Different effects of rule-guided inference on major SGs and minor SGs

To further investigate the decrease in Top-1 performance at a training-data ratio of 0.8, we compared the Top-1 prediction obtained from diversity-enhanced retrieval with the final ranking produced by DyRIS. As shown in **Table 7**, the retrieval-only Top-1 accuracy for these two major SGs was higher than the final DyRIS Top-1 accuracy at both training-data ratios. Specifically, retrieval-only Top-1 accuracy was 0.8267 and 0.8200 at training-data ratios of 0.5 and 0.8, respectively, whereas the corresponding final DyRIS Top-1 accuracy was 0.6667 and 0.6067. The gap therefore increased from 16.00 percentage points at a training-data ratio of 0.5 to 21.33 percentage points at a training-data ratio of 0.8.

This result indicates that, for major SGs, diversity-enhanced retrieval can already provide strong evidence for the correct Top-1 prediction. However, the final ranking step based on rule-guided inference, which jointly considers quantitative indicators, can sometimes weaken this strong retrieval evidence and demote the correct major SG from the Top-1 position. In the high-data regime, the retrieval-only prediction for major SGs remains strong, but the performance gap between retrieval-only prediction and final DyRIS ranking becomes larger, indicating that the ranking step has a stronger adverse effect on major SG Top-1 prediction at a training-data ratio of 0.8.

In contrast, for minor SGs, the final DyRIS ranking improved Top-1 performance compared with diversity-enhanced retrieval alone as shown in **Table 7**. Minor-SG Top-1 accuracy increased from 0.3111 to 0.3762 at a training-data ratio of 0.5 and from 0.2804 to 0.3534 at a training-data ratio of 0.8. These correspond to improvements of 6.51 and 7.30 percentage points, respectively. The lower retrieval-only Minor-SG Top-1 accuracy suggests that Top-1 prediction based only on the nearest retrieved example is sensitive for minor SGs. The improvement by DyRIS was larger at a training-data ratio of 0.8 than at 0.5, suggesting that the benefit of rule-guided inference for minor SGs becomes more pronounced in the higher-data regime. Because minor SGs have fewer training samples and can be located near structurally similar major SGs or other minor SGs in the embedding space, relying only on a single nearest retrieved example can lead to unstable Top-1 predictions.

These results show that the effect of the final DyRIS ranking becomes more class-dependent as the training-data ratio increases. For major SGs, the final ranking step can reduce Top-1 performance relative to diversity-enhanced retrieval alone. For minor SGs, however, the same rule-guided inference improves Top-1 prediction by integrating candidate-level quantitative evidence and structural prior information. This indicates that the quantitative-indicator-based rule-guided inference in DyRIS is effective for the main objective of improving Minor-SG Top-1 prediction, while the same ranking strategy can have an opposite effect on major SGs. The following section further analyzes how rule-guided inference affects Top-1 prediction for major SGs and minor SGs based on the quantitative indicators.

**Table 7.** Comparison between retrieval-only Top-1 prediction and the final DyRIS Top-1 prediction for major SGs and minor SGs. Retrieval-only performance was evaluated using the first-ranked candidate obtained from diversity-enhanced retrieval before rule-guided inference. DyRIS performance represents the final Top-1 prediction after rule-guided inference. The difference is reported as DyRIS minus retrieval-only performance in percentage points. The analysis was performed using the same five data splits used in Section 3.1.2, and the reported values represent averages over the five splits.

| SG type | Training ratio | Retrieval only | DyRIS | Differences |
|---|---|---|---|---|
| Major SGs | 0.5 | 0.8267 | 0.6667 | -16.00pp |
| | 0.8 | 0.8200 | 0.6067 | -21.33pp |
| Minor SGs | 0.5 | 0.3111 | 0.3762 | +6.51pp |
| | 0.8 | 0.2804 | 0.3534 | +7.30pp |

### 3.3.3 Class-dependent analysis of quantitative-indicator-based Top-1 ranking

The results in Section 3.3.2 show that the final rule-guided ranking of DyRIS affects major SGs and minor SGs in opposite ways. Correct major SGs can be demoted from the Top-1 position, whereas Top-1 prediction for minor SGs can be improved. To understand the source of this class-dependent behavior, we analyzed how the quantitative indicators provided in the prompt were reflected in the final Top-1 ranking. This analysis was performed using the five experimental splits described in Section 3.1.2.

First, we focused on misclassified cases in which the ground-truth SG was SG 14 or SG 225, but DyRIS failed to predict it as Top-1. For each case, we compared the SG selected by DyRIS as the final Top-1 prediction with the ground-truth SG using Retrieval rank, Combined score, Global fit, $z_{abs_mean}(s)$, $z_{max_abs}(s)$ and $z_{best_count}(s)$. As shown in **Table 8**, the ground-truth SG more frequently had a better Retrieval rank, indicating that it was often closer to the query composition in the retrieved example list. However, for the remaining quantitative indicators, including Combined score, Global fit, and feature-wise consistency, the SG selected by DyRIS as Top-1 more frequently showed better indicator values than the ground-truth SG. This tendency became more pronounced at a training-data ratio of 0.8, where other than Retrieval rank most quantitative indicators more frequently showed better values for the DyRIS-selected Top-1 SG over the ground-truth SG.

This result indicates that the decrease in Top-1 performance for major SGs did not arise from arbitrary behavior of the LLM agent. Instead, it originated from conflicts between Retrieval rank and other quantitative evidence. In other words, for major SGs, the nearest retrieved example often already provided strong evidence for the correct SG, but the final rule-guided ranking could demote the ground-truth SG when other indicators showed better indicator values for a different candidate.

**Table 8.** Quantitative-indicator comparison between the DyRIS-predicted Top-1 SG and the ground-truth SG in misclassified major-SG cases. The analysis was performed for samples whose ground-truth SG was SG 14 or SG 225 but whose final DyRIS Top-1 prediction was incorrect. For each metric, the table reports the number of cases in which either the DyRIS-predicted Top-1 SG or the ground-truth SG showed the better indicator value, along with the number of equal cases.

| | **Training-data ratio = 0.5** | | | **Training-data ratio = 0.8** | | |
|---|---|---|---|---|---|---|
| Metric | DyRIS Top-1 SG better | Ground-truth SG better | Equal | DyRIS Top-1 SG better | Ground-truth SG better | Equal |
| Retrieval rank | 30 | 67 | 0 | 39 | 77 | 0 |
| Combined score | 70 | 27 | 0 | 94 | 22 | 0 |
| Global fit | 82 | 15 | 0 | 98 | 18 | 0 |
| $z_{abs_mean}(S)$ | 78 | 19 | 0 | 94 | 22 | 0 |
| $z_{max_abs}(S)$ | 75 | 22 | 0 | 87 | 29 | 0 |
| $z_{best_count}(S)$ | 71 | 8 | 18 | 87 | 6 | 23 |

In contrast, the opposite trend was observed for minor SGs. For samples whose ground-truth labels were minor SGs, we compared cases in which the retrieval-only Top-1 prediction was incorrect but the final DyRIS Top-1 prediction was correct with cases in which the retrieval-only Top-1 prediction was correct but the final DyRIS Top-1 prediction was incorrect. As shown in **Table 9**, DyRIS corrected retrieval-only Top-1 errors more frequently than it introduced errors by changing correct retrieval-only predictions. At a training-data ratio of 0.5, DyRIS corrected 160 retrieval-only errors, whereas 67 retrieval-only correct predictions became incorrect after DyRIS ranking, giving a net gain of 93 cases. At a training-data ratio of 0.8, the corresponding numbers were 124 and 57, resulting in a net gain of 67 cases.

A detailed comparison of the quantitative indicators for these correction cases is provided in **Section S7, SI.** The results show that the ground-truth SG selected by DyRIS more frequently showed better indicator values than the retrieval-only Top-1 candidate across most quantitative indicators. This indicates that DyRIS often promoted the correct SG to the Top-1 position for minor SGs by using the quantitative indicators provided in the prompt, rather than simply following the retrieval order.

Overall, the evidence-integrated ranking of DyRIS had opposite effects on major SGs and minor SGs. For major SGs, retrieval proximity already provided strong evidence for the correct SG, so incorporating additional quantitative indicators could demote the correct SG from the Top-1 position in some cases. For minor SGs, however, retrieval-only Top-1 prediction was less stable, and integrating quantitative indicators effectively corrected such errors. These results suggest that the LLM agent reflected the structured evidence provided in the prompt during ranking. They also indicate that a class-adaptive ranking strategy may be needed: retrieval proximity should be preserved more strongly for major SGs, whereas evidence-integrated ranking should be applied more actively for minor SGs.

**Table 9.** Comparison of retrieval-only and final DyRIS Top-1 predictions for minor SGs. The table reports the number of cases in which diversity-enhanced retrieval produced an incorrect Top-1 prediction but DyRIS corrected it, and the number of cases in which diversity-enhanced retrieval produced a correct Top-1 prediction but DyRIS changed it to an incorrect prediction. The net value is calculated as the number of corrected cases minus the number of newly introduced errors.

| Training ratio | Retrieval wrong and DyRIS correct | DyRIS wrong and Retrieval correct | Net |
|---|---|---|---|
| 0.5 | 160 | 67 | +93 |
| 0.8 | 124 | 57 | +67 |

#### 3.3.4 Complementarity between DyRIS and PyCaret-based model

The preceding case-level analyses focused on the relationship between diversity-enhanced retrieval and evidence-integrated ranking within DyRIS. In this section, we examine the complementarity between DyRIS and a data-driven baseline model. As discussed in Section 3.1.2, under the high-data condition corresponding to a training-data ratio of 0.8, the PyCaret-based model showed the strongest Top-3 performance among the baseline models. This indicates that a supervised classifier trained with sufficient data is effective at broadly including the correct SG within the Top-3 candidates. In contrast, DyRIS showed stronger performance in Top-1 ranking and macro-F1 score for minor SGs under the same condition. These results suggest that the strengths of the two models are complementary: the PyCaret-based model is effective at including the correct SG within the Top-3 candidates, whereas DyRIS is effective at evidence-based ranking and correction for minor SGs. Combining these complementary strengths is expected to improve SG prediction more broadly, because including the correct SG within the Top-3 candidate set provides a better starting point for subsequent Top-1 ranking.

To examine this possibility, we performed a preliminary analysis using a hybrid approach that heuristically combines the predictions of DyRIS and the PyCaret-based model. The hybrid approach determines the final Top-3 SG candidates and their ranking by applying heuristic rules to the Top-3 predictions obtained from the two models. The detailed heuristic rules are provided in **Section S8, SI**. The performance of the hybrid approach was evaluated using the same training-data ratio of 0.8 and the same five data splits used in Section 3.1.2.

As shown in **Figure 6**, the hybrid approach improved all evaluated metrics compared with DyRIS and PyCaret-based model. For Top-3 accuracy, the hybrid approach slightly exceeded the PyCaret-based model while improving over DyRIS. At the same time, it preserved the advantage of DyRIS in minor-SG Top-1 and macro-F1 performance, and even further improved these metrics. These results suggest that combining

DyRIS with the PyCaret-based model can mitigate the limitations of DyRIS in the high-data regime while retaining its strength in predicting minor SGs.

However, the current hybrid approach is a preliminary strategy based on heuristic rules. Future work should develop a more systematic hybrid inference strategy that jointly considers model confidence, rank agreement, class type, and evidence consistency.

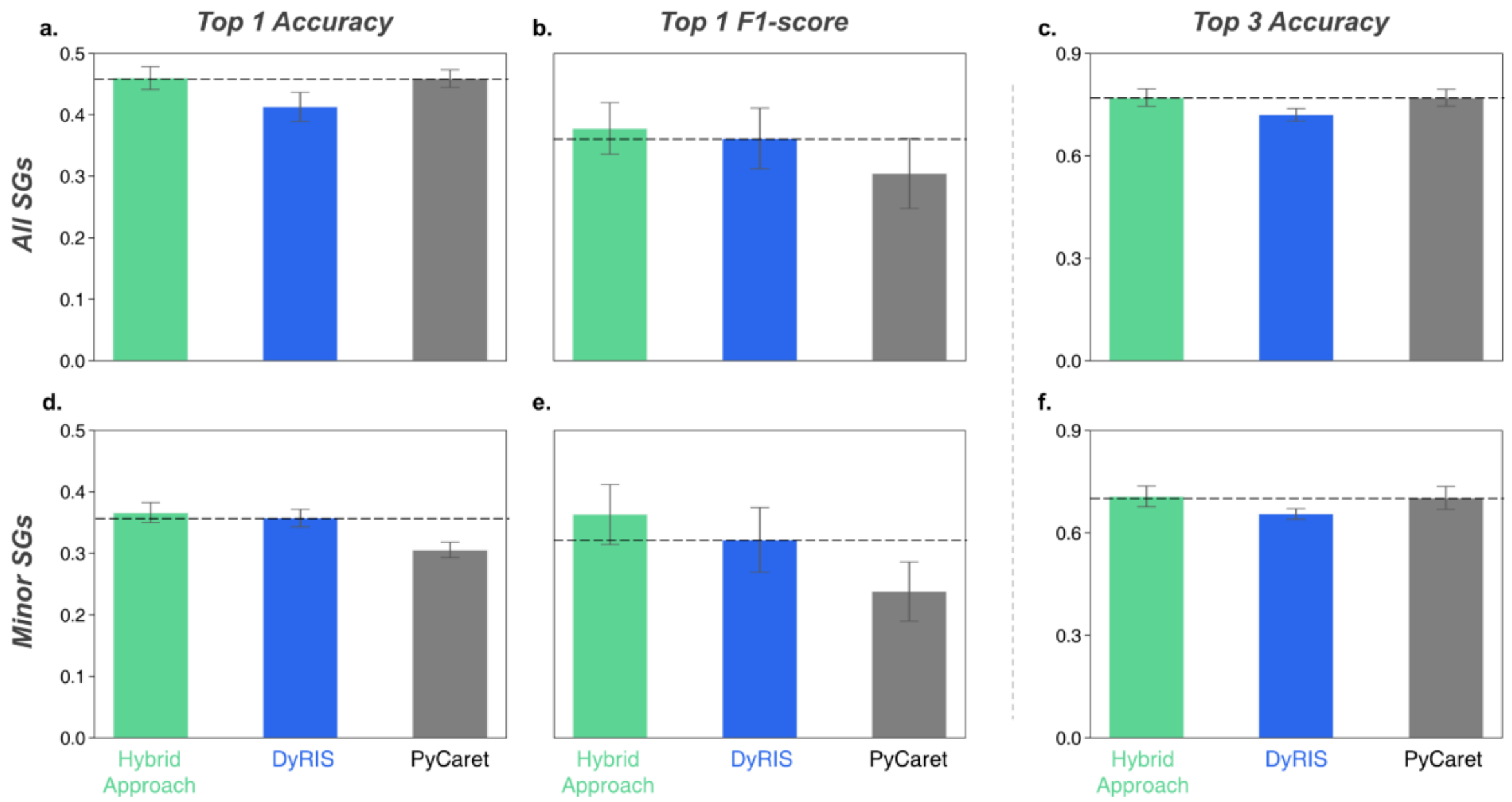

**Figure 6.** Performance comparison of the hybrid approach, DyRIS, and the PyCaret-based model at a training-data ratio of 0.8. (a–c) Overall Top-1 accuracy, Top-1 macro-F1 score, and Top-3 accuracy evaluated on all SG classes. (d–f) Corresponding metrics evaluated on the minor-SG subset. Error bars represent the standard deviation over five random train/test splits. The black dashed line in each panel indicates the second-highest mean performance among the three models, highlighting whether the best-performing model provides an improvement over the next-best model.

# 4. Conclusion

## 4.1 Summary

This study proposed DyRIS, an LLM-agent-based framework for predicting ranked SG candidates corresponding to stable DP structures from a given DP composition. DyRIS combines diversity-enhanced dynamic few-shot retrieval with rule-guided inference based on $B/B'$ cation ordering and quantitative indicators. This design allows DyRIS to select in-context examples that are close to the query composition in the embedding space while maintaining SG-label diversity so that major SGs do not dominate the prompt. The LLM agent then uses crystallographic prior information and candidate-level quantitative evidence to refine and ranks the final Top-3 SG candidates.

The results show that DyRIS provides competitive overall performance compared with data-driven baseline models while improving prediction performance for minor SGs. At a training-data ratio of 0.5, DyRIS achieved performance comparable to the strongest baselines in Overall Top-1 accuracy and Overall Top-3 accuracy, while achieving the best performance in Overall Top-1 macro-F1 score and all Minor-SG metrics. This indicates that DyRIS does not improve performance simply by favoring major SGs, but provides more balanced SG prediction under severe class imbalance.

The ablation study further confirmed that each component of DyRIS contributes to the final prediction performance. Diversity-enhanced dynamic few-shot retrieval played the primary role in narrowing the possible SG candidate space to plausible candidates. Quantitative indicators improved candidate selection and ranking by providing evidence on the local SG density around the query composition and the distributional agreement between the query composition and candidate SGs. The major-SG bias-control mitigates majority-class over-ranking and supports minor-SG prediction. $B/B'$ ordering information contributed as a soft structural prior, with a more evident effect in the high-data regime. In addition, when the final rule-guided inference step was replaced with conventional ML-based re-ranking models, the ML models did not out-perform DyRIS in any of the evaluated metrics, with particularly large gaps in Minor-SG Top-1 metrics. These results suggest that the strength of DyRIS is associated with the inference process that integrates retrieval evidence, quantitative evidence, and crystallographic prior information, rather than with a single rule or a single feature.

### 4.2 Limitations and future work

Despite these advantages, DyRIS has several limitations. The main limitation is that increasing the amount of training data does not consistently improve all performance metrics. Several Top-1 metrics decreased when the training-data ratio increased from 0.5 to 0.8. Further analysis showed that this limitation was mainly associated with the final ranking step, where the correct SG must be selected as the final Top-1 prediction from the retrieved candidate set.

The final rule-guided ranking showed different effects depending on class type. For major SGs, diversity-enhanced dynamic few-shot retrieval alone provided strong Top-1 evidence, but the final rule-guided ranking could demote the correct major SG from the Top-1 position when other quantitative indicators supported a different candidate. In contrast, for minor SGs, retrieval-only Top-1 prediction was relatively unstable, and the same rule-guided inference often corrected retrieval errors by integrating quantitative indicators and ordering information. This suggests that a single ranking strategy may not be optimal for both major SGs and minor SGs. Future work should therefore develop a class-adaptive ranking strategy that preserves retrieval proximity more strongly for major SGs while applying evidence-integrated ranking more actively for minor SGs.

Another important direction is to develop a more systematic hybrid inference framework. In the high-data regime, the PyCaret-based model showed strong Top-3 performance, whereas DyRIS showed stronger Top-1 ranking and macro-F1 performance for minor SGs. The preliminary hybrid approach demonstrated that combining the two models can improve all evaluated metrics compared with DyRIS alone. However, the current hybrid strategy is based on heuristic rules. Future studies should develop a principled hybrid inference method that jointly considers model confidence, rank agreement, class type, and evidence consistency. Such a framework could preserve the strength of DyRIS in minor-SG prediction while improving the stability of overall SG prediction in high-data regimes.

**Supporting Information**

# Predicting Space Groups of Double Perovskites by LLM with Dynamic Few-Shot Learning

Jongwon Park [1,†], Inhyo Lee[1,†], Junhyeong Lee[1], and Seunghwa Ryu[1,2,3,*]

**Affiliations**

[1]Department of Mechanical Engineering, Korea Advanced Institute of Science and Technology (KAIST), 291 Daehak-ro, Yuseong-gu, Daejeon 34141, Republic of Korea

[2]KAIST InnoCORE PRISM-AI Center, Korea Advanced Institute of Science and Technology (KAIST), Daejeon, 34141, Republic of Korea

[3]Department of AX, Korea Advanced Institute of Science and Technology (KAIST), 291 Daehak-ro, Yuseong-gu, Daejeon 34141, Republic of Korea

[*]Corresponding author: ryush@kaist.ac.kr

# 1. Composition-based feature construction

## 1.1 Oxidation-state assignment

The construction of DP-specific features requires consistent assignment of oxidation states because Shannon ionic radii depend on both oxidation state and coordination environment. In this study, oxidation states were assigned using a three-step procedure based primarily on charge neutrality. First, a rule-based search was performed by combining representative oxidation-state candidates for each element and selecting combinations that satisfy charge neutrality. If this procedure failed, representative anion oxidation states for O, F, Cl, Br, and I were assumed, and possible cation oxidation-state combinations were enumerated to identify charge-neutral solutions. If both rule-based procedures failed because of rare elemental combinations or compositional complexity, an LLM was used as an auxiliary solver to propose chemically plausible oxidation-state mappings. The LLM-generated candidates were accepted only after verifying charge neutrality and stoichiometric consistency.

## 1.2 Site assignment and Shannon ionic radius lookup

After oxidation-state assignment, the elements in each DP composition were assigned to the $A$, $B$, $B'$, and $X$ crystallographic sites. Shannon ionic radii were then obtained from a lookup table using the assigned oxidation state and coordination number. Because the Shannon radius depends on coordination environment, we defined site-specific coordination-number priorities based on typical perovskite environments. The $A$ site was treated as a large-cation site with high coordination, whereas the $B$ and $B'$ sites were treated as octahedrally coordinated sites. The $X$ site was assigned using an anion-specific coordination-number priority. If an exact match for both oxidation state and coordination number was unavailable, the radius with the nearest available coordination number was used. If no radius was available under this criterion, the radius with the nearest oxidation state was used to reduce missing values.

**Table S1.** Coordination-number priority used for Shannon ionic radius lookup.

| **Site** | **Coordination-number priority** |
|---|---|
| *A* site | 12 → 9 → 8 → 10 → 11 |
| *B*/*B*′ site | 6 → 5 → 4 → 7 |
| *X* site | 6 → 4 → 8 |

### 1.3 DP-specific and *B*-site-derived feature calculation

The extracted oxidation states and Shannon ionic radii were used to construct the DP-specific features described in the main text. The effective *B*-site radius was defined as the arithmetic mean of the *B* and *B*′- site ionic radii and was used to calculate the tolerance factor and octahedral factor. In addition, the *B* and *B*′-site ionic radii were used to construct *B*-site-derived features, including the radius difference and radius ratio between the two *B*-site cations.

### 1.4 Magpie feature construction

Magpie features were constructed to represent composition-level chemical statistics. We used the open-source Python library matminer and its ElementProperty featurizer with the "magpie" preset. For each composition, elemental properties were retrieved and summarized using statistical descriptors such as mean, deviation, minimum, maximum, and range. These descriptors were included to capture global compositional trends that may not be fully represented by the DP-specific, site-level, and *B*-site-derived features.

### 1.5 Final feature vector and embedding construction

The final feature vector for each DP composition was constructed by concatenating the DP-specific features, site features, *B*-site-derived features, and Magpie features. Each feature had dimensionalities of 10, 16, 3, and 132, respectively, resulting in a 161-dimensional feature vector for each composition. This concatenated vector was used as the base representation for constructing the embedding space in diversity-enhanced dynamic few-shot retrieval. In the main DyRIS framework, feature-block weighting was further applied to this vector to account for the different contributions of each feature group to SG prediction.

## 2. Feature-block weighting for embedding construction

To construct a more informative embedding space for diversity-enhanced dynamic few-shot retrieval, we introduced feature-block weighting. The final composition-based feature vector consists of four feature blocks: site features, DP-specific features, *B*-site-derived features, and Magpie features. We assigned a block-level weight to each feature group, denoted as *a, b, c,* and *d*, respectively. Here, *a* corresponds to the site features, *b* to the DP-specific features, *c* to the *B*-site-derived features, and *d* to the Magpie features.

The weighted embedding vector was constructed by scaling each feature block before concatenation. The weight combination $(a, b, c, d)$ was constrained to satisfy $a + b + c + d = 1$. To identify an effective weighting scheme, we performed an exhaustive grid search over the simplex using a step size of 0.05. Because the number of possible combinations was limited and the evaluation cost for each candidate weight set was manageable, direct grid search was used instead of a continuous optimization algorithm.

Each weight combination was evaluated using a leave-one-out procedure within the training data. Specifically, one training sample was treated as a query composition, and the remaining training samples were used as the retrieval pool. Diversity-enhanced retrieval was then applied to select in-context examples for the query. This process was repeated for all training samples. Retrieval performance was evaluated using Top-1, Top-3, and Top-5 accuracy, where each metric indicates whether the true SG of the query composition was included within the corresponding retrieved candidate set. The retrieval performance obtained before and after feature-block weighting is summarized in **Table S2.**

**Table S2.** Retrieval performance before and after feature-block weighting. The evaluation was performed using a leave-one-out procedure within the training data at a training-data ratio of 0.5. Overall Top-1, Top-3, and Top-5 accuracies indicate whether the true SG of the query composition was included within the corresponding retrieved candidate set. Values are reported as mean ± standard deviation over five random train/test splits.

| Metric | Before weighting (mean ± std) | After weighting (mean ± std) |
|---|---|---|
| Overall Top-1 accuracy | 0.379 ± 0.026 | 0.402 ± 0.012 |
| Overall Top-3 accuracy | 0.653 ± 0.010 | 0.675 ± 0.007 |
| Overall Top-5 accuracy | 0.783 ± 0.012 | 0.802 ± 0.013 |

## 3. *B/B′* ordering surrogate model

The collected DP dataset did not include explicit *B/B′* cation-ordering labels. Therefore, ordering labels were assigned from the available structure files, including CIF and POSCAR files. We used pymatgen to analyze the local coordination environments of *B* and *B′*-site cations and classified the *B/B′* ordering type into four categories: rock-salt, columnar, layered, and other ordering. The other ordering class was assigned when the local *B/B′* arrangement could not be classified into one of the three ordered patterns.

Using these structure-derived ordering labels, we trained a PyCaret-based ordering surrogate model. The input to this model was the composition-based feature vector described in Section 2.2 of the main text, including DP-specific features, site features, *B*-site-derived features, and Magpie features. The surrogate model predicts the class probabilities for the four *B/B′* ordering types.

The ordering surrogate model was evaluated on the test data corresponding to training-data ratios of 0.5 and 0.8 used in the main experiments. Because DyRIS uses the predicted ordering information as a soft structural prior rather than as a deterministic ordering label, the usefulness of the ordering surrogate model should not be assessed only by whether the most probable ordering type exactly matches the structure-derived label. Instead, a more relevant criterion is whether the SG set compatible with the predicted ordering information contains the ground-truth SG. We refer to this metric as **ordering-compatible coverage.** As shown in Table S3, the coverage is enough high in both training ratios. This indicates that the predicted ordering rarely excludes the correct SG, confirming that it provides a reliable structural prior for DyRIS.

**Table S3**. Ordering-compatible coverage of the *B/B′* ordering surrogate model. Coverage indicates the fraction of query compositions for which the ground-truth SG is included in the SG set compatible with the predicted ordering information. Values are reported as mean ± standard deviation over five random train/test splits.

| Metric | Training ratio= 0.5 | Training ratio= 0.8 |
|---|---|---|
| Overall Top-1 accuracy | 0.9675 ± 0.0156 | 0.9689 ± 0.0206 |

## 4. Prompt and query template for rule-guided LLM inference

The following prompt template was used for rule-guided LLM inference in DyRIS. Auxiliary terms that are not defined in the main text, including contradiction_count, SG_knn, and SG_fit, are defined within the prompt. The full prompt template is provided below:

[prompt]

[Role]

You are an expert in chemistry and materials science, with deep knowledge of double perovskites A2BB'X6, B/B' ordering, octahedral tilting, structural distortion, and crystallographic space groups.

[Task]

Predict the Top-3 most plausible space-group numbers for the target double-perovskite composition.

The prediction should correspond to the lowest-E_hull structure.

Use your pre-trained materials-science knowledge to interpret the provided evidence, but prioritize the query-specific data and the allowed candidate set.

[Allowed candidate space groups]

Choose only from the following 19 candidates:

[1, 2, 7, 11, 12, 14, 26, 31, 34, 87, 123, 139, 146, 148, 164, 166, 194, 201, 225]

[Important data context]

The training data are imbalanced. SG 14 and SG 225 are majority classes.

[Definitions]

1. z value

z = (query_feature - SG_training_mean) / SG_training_std

A smaller absolute z value means that the query feature is more typical of the candidate SG distribution.

2. combined_score

combined_score measures the local SG support around the query composition in the embedding space.

A larger combined_score means stronger local support.

combined_score is the primary anchor for determining Top-3 inclusion.

3. global_fit

global_fit = sqrt(z_tf^2 + z_of^2 + z_rA^2 + z_ratio^2 + z_Delta_rB^2 + z_Delta_EN^2)

A smaller global_fit means that the query is more typical of the SG-wise training distribution.

4. feature-wise z consistency

feature-wise z consistency means that the absolute z values of the features are generally small and that there is no excessive mismatch in any specific feature. Interpret it as follows:

- z_abs_mean: smaller is better.

- z_max_abs: smaller is better.

- z_best_count: larger is better.

5. ordering probability

Ordering probabilities are soft evidence for the B/B' ordering tendency.

Because rare ordering types are uncommon, the rare ordering probabilities given in the query may be numerically small.

A rare-ordering signal should not be ignored solely because its raw probability is lower than the rock-salt probability.

[Compatible ordering: Ordering type → SG sets]

broken: {12, 164, 166, 194}

layered: {11, 26, 123}

columnar: {11, 34}

rock-salt: {1, 2, 7, 12, 14, 31, 87, 139, 146, 148, 201, 225}

[Top-3 inclusion policy]

1. Combined-score anchor

Start from the three candidates with the highest combined_score.

This default Top-3 is the primary candidate set.

2. KNN-consensus retention

Also consider the three earliest KNN retrieved candidates as independent local-analogy evidence.

Candidates that appear in both the combined-score Top-3 and the KNN Top-3 should be strongly retained in the final Top-3.

3. Boundary candidate comparison

If the combined-score Top-3 and the KNN Top-3 disagree, compare candidates that appear in only one of the two groups.

Prefer candidates supported by several converging signals:

- compatibility with the dominant ordering or relevant rare ordering

- better global_fit

- better feature-wise z consistency

Do not replace a candidate supported by both combined_score evidence and KNN evidence unless several other signals consistently contradict it.

4. Role of ordering

Ordering compatibility is important for Top-3 inclusion.

When the other evidence is comparable, prefer candidates compatible with the dominant ordering.

A rare-ordering-compatible candidate may be retained or promoted when a rare-ordering signal exists and the candidate also has local support from combined_score or KNN retrieval.

[Ranking policy]

After selecting the final Top-3 set, determine the final ranking using the following evidence:

- KNN ranking: earlier is better.

- combined_score: larger is better.

- global_fit: smaller is better.

- feature-wise z consistency: z_abs_mean and z_max_abs are better when smaller, and z_best_count is better when larger.

- Because SG 14 and SG 225 are overrepresented in the imbalanced training data, they should be selected as Top-1 only when they are clearly better than the other space groups across multiple indicators.

[Output requirement]

Do not output the reasoning process.

Return only one line:

[SG1, SG2, SG3]

[query_text]

[Query]

[Target composition]

Formula: Sr2TiIrO6

Sites:

A  : ox = +2, r = 1.440

B  : ox = +4, r = 0.605

B' : ox = +4, r = 0.625

X  : ox = -2, r = 1.400

Target features:

tf = 0.997

of = 0.439

r_A = 1.440

radius_ratio_B = 1.033

Delta r_B = 0.020

Delta EN_B = 0.660

[Ordering evidence]

Ordering | predicted_p | compatible_candidates

broken   | 0.01 | 12

columnar | 0.00 | none

layered  | 0.00 | none

rock-salt | 0.99 | 12, 2, 225, 14, 146

[Candidate-set summary]

majority_candidates_present: 225, 14

non_major_candidates_present: 12, 2, 146

rare_ordering_compatible_candidates_present:

  broken: 12

  columnar: none

  layered: none

rock_salt_compatible_candidates_present: 12, 2, 225, 14, 146

[Retrieved examples]

Example 1:

SG = 12

formula = Sr2VOsO6

feature_difference_from_query:

d_tf = -0.005

d_of = +0.007

d_rA = +0.000

d_ratio = -0.053

d_Delta_rB = -0.030

d_Delta_EN = +0.090

Example 2:

SG = 2

formula = Sr2MnOsO6

feature_difference_from_query:

d_tf = -0.018

d_of = +0.025

d_rA = +0.000

d_ratio = -0.156

d_Delta_rB = -0.080

d_Delta_EN = +0.010

Example 3:

SG = 225

formula = Sr2VWO6

feature_difference_from_query:

d_tf = -0.007

d_of = +0.011

d_rA = +0.000

d_ratio = -0.189

d_Delta_rB = -0.100

d_Delta_EN = -0.070

Example 4:

SG = 14

formula = Ca2TiIrO6

feature_difference_from_query:

d_tf = +0.035

d_of = +0.000

d_rA = +0.100

d_ratio = +0.000

d_Delta_rB = +0.000

d_Delta_EN = +0.000

Example 5:

SG = 146

formula = Sr2CoOsO6

feature_difference_from_query:

d_tf = -0.018

d_of = +0.025

d_rA = +0.000

d_ratio = -0.156

d_Delta_rB = -0.080

d_Delta_EN = +0.340

[Candidate evidence]

Candidates are sorted by combined_score.

SG | knn_order | combined_score | global_fit | compatible_ordering

12 | 1 | 0.0995 | 1.467 | broken, rock-salt

2 | 2 | 0.0995 | 1.240 | rock-salt

14 | 4 | 0.0207 | 2.059 | rock-salt

146 | 5 | 0.0087 | 1.366 | rock-salt

225 | 3 | 0.0023 | 2.271 | rock-salt

[KNN and combined-score grouping]

knn_top3_candidates: 12, 2, 225

combined_score_top3_candidates: 12, 2, 14

consensus_candidates: 12, 2

combined_score_only_candidates: 14

knn_only_candidates: 225

[Top-3 boundary comparison set]

candidates_to_compare_for_top3: 12, 2, 14, 225

[Majority-crowding summary]

majority_candidates_in_combined_score_top3: 14

majority_candidates_in_knn_top3: 225

non_major_candidates_in_boundary_set: 12, 2

[Feature-wise z values]

| SG | z_tf | z_of | z_rA | z_ratio | z_Delta_rB | z_Delta_EN |
|---|---|---|---|---|---|---|
| 12 | -0.032 | -0.450 | -0.815 | +0.211 | -0.963 | +0.557 |
| 2 | +0.509 | -0.558 | -0.092 | +0.197 | -0.722 | +0.631 |
| 14 | +1.314 | -0.864 | +0.715 | +0.438 | -0.967 | +0.355 |
| 146 | +0.298 | -0.801 | -0.572 | +0.075 | -0.882 | +0.155 |
| 225 | +0.813 | -0.830 | -1.549 | +0.228 | -1.143 | +0.216 |

[Feature-wise z consistency summary]

| SG | z_abs_mean | z_max_abs | z_best_count |
|---|---|---|---|
| 12 | 0.505 | 0.963 | 2 |
| 2 | 0.451 | 0.722 | 2 |

14 | 0.775 | 1.314 | 0

146 | 0.464 | 0.882 | 2

225 | 0.797 | 1.549 | 0

## 5. Performance of retrieval-only prediction

To examine whether the retrieval stage alone is sufficient for SG prediction, we evaluated a retrieval-only setting at a training-data ratio of 0.5. This setting used the same train/test splits as those used in Section 3.1.1. The weighted embedding and diversity-enhanced dynamic few-shot retrieval procedure were kept identical to DyRIS, but the subsequent rule-guided LLM-based SG inference step was removed. The retrieved SG candidates were directly used as ranked predictions.

**Table S4** compares the retrieval-only performance with the full DyRIS performance. Retrieval-only prediction showed moderate performance, indicating that the weighted diversity-enhanced retrieval stage can identify plausible SG candidates. However, its performance was consistently lower than that of full DyRIS across all evaluated metrics. These results indicate that retrieval alone is insufficient for final SG prediction. Although diversity-enhanced retrieval narrows the candidate space to plausible SGs, the rule-guided LLM-based inference step further improves candidate selection and ranking.

**Table S4.** Retrieval-only performance compared with full DyRIS at a training-data ratio of 0.5. All values are reported as mean ± standard deviation over five random train/test splits. Retrieval-only prediction uses the weighted diversity-enhanced retrieval results directly without rule-guided LLM-based inference.

| Metric | Overall Top-1 accuracy | Overall Top-1 macro-F1 score | Overall Top-3 accuracy | Minor-SG Top-1 accuracy | Minor-SG Top-1 macro-F1 score | Minor-SG Top-3 accuracy |
|---|---|---|---|---|---|---|
| Full DyRIS | 0.4266 ± 0.0230 | 0.3846 ± 0.0116 | 0.6994 ± 0.0188 | 0.3762 ± 0.0226 | 0.3388 ± 0.0061 | 0.6658 ± 0.0242 |
| Retrieval-only | 0.3904 ± 0.0209 | 0.3370 ± 0.0214 | 0.6496 ± 0.0293 | 0.3028 ± 0.0196 | 0.2910 ± 0.0207 | 0.5886 ± 0.0277 |

## 6. Effect and actionability of ordering information

### 6.1 Ordering ablation across training-data ratios

Because the effect of ordering information was limited and mixed at a training-data ratio of 0.5, we further examined whether ordering information should be retained in the DyRIS prompt and query. To this end, we compared full DyRIS with DyRIS without ordering information at training-data ratios of 0.5 and 0.8.

**Table S5.** Ablation analysis of ordering information at training-data ratios of 0.5 and 0.8. Full DyRIS is compared with DyRIS without ordering information. The change, Δ, is calculated as the performance of DyRIS without ordering information minus that of full DyRIS and is reported in percentage points.

| Metric | Full DyRIS, 0.5 | w/o ordering, 0.5 | Δ pp | Full DyRIS, 0.8 | w/o ordering, 0.8 | Δ pp |
|---|---|---|---|---|---|---|
| Overall Top-1 acc | 0.4266 | 0.4279 | 0.13 | 0.4188 | 0.396 | -2.28 |
| Overall Top-1 macro-F1 score | 0.3846 | 0.3885 | 0.39 | 0.3694 | 0.3504 | -1.9 |
| Overall Top-3 acc | 0.6994 | 0.6713 | -2.81 | 0.7352 | 0.7278 | -0.74 |
| Minor-SG Top-1 acc | 0.3762 | 0.3693 | -0.69 | 0.361 | 0.3186 | -4.24 |
| Minor-SG Top-1 macro-F1 score | 0.3388 | 0.3831 | 4.43 | 0.3312 | 0.3126 | -1.86 |
| Minor-SG Top-3 acc | 0.6658 | 0.6713 | 0.55 | 0.6824 | 0.6728 | -0.96 |

The mixed effect at a training-data ratio of 0.5 suggests that ordering information is not always decisive when retrieval evidence and quantitative indicators already provide sufficient support for candidate selection. However, at a training-data ratio of 0.8, removing ordering information led to consistent performance decreases across all metrics. This indicates that ordering information becomes more useful when the retrieved candidate set contains more competing SG candidates, because ordering compatibility can help distinguish crystallographically plausible candidates among otherwise similar candidates.

### 6.2 Selectivity and actionability within the retrieved Top-5 candidate set

We next analyzed the selectivity of ordering information within the retrieved Top-5 candidate set. Unlike ordering-compatible coverage, which measures whether the ground-truth SG is included in the ordering-compatible SG set, selectivity measures whether ordering information can distinguish among the retrieved candidates. In this analysis, active ordering types were defined as the ordering types whose predicted probabilities were provided to the LLM as ordering evidence. The active ordering-compatible SG set was then constructed as the union of the SG sets compatible with these active ordering types.

To quantify the selectivity of ordering information, we evaluated several candidate-set-level metrics. The rare-ordering active rate denotes the fraction of samples for which at least one rare ordering type was active. The compatible SG union size denotes the number of SGs included in the union of active ordering-compatible SG sets; a smaller value indicates that the ordering prior provides a narrower compatible SG set. The Top-5 compatible candidate count denotes the average number of retrieved Top-5 SG candidates that are compatible with the active ordering information. The all-Top-5-compatible rate denotes the fraction of samples for which all retrieved Top-5 candidates are ordering-compatible. When this value is high, ordering information has limited discriminative power because it cannot distinguish among the retrieved candidates.

We further defined the actionable-case rate to measure how often ordering information can actually affect candidate selection. A case was considered actionable when three conditions were satisfied: the ground-truth SG was included in the retrieved Top-5 candidates, the ground-truth SG was included in the active ordering-compatible SG set, and not all retrieved Top-5 candidates were ordering-compatible. In such cases, ordering information has the potential to retain a compatible correct candidate while weakly filtering or demoting incompatible candidates. The minor-SG actionable-case rate was calculated using the same criterion but only for samples whose ground-truth labels were minor SGs.

**Table S6** Selectivity and actionability of ordering information within the retrieved Top-5 candidate set. The rare-ordering active rate indicates the fraction of samples for which at least one rare ordering type was active. The compatible SG union size is the number of SGs included in the union of active ordering-compatible SG sets. The Top-5 compatible candidate count is the average number of retrieved Top-5 candidates compatible with the active ordering information. The all-Top-5-compatible rate indicates the fraction of samples for which all retrieved Top-5 candidates were ordering-compatible. The actionable-case rate indicates the fraction of samples for which ordering information could potentially distinguish compatible and incompatible candidates while retaining the ground-truth SG. Values are reported as mean ± standard deviation over five random train/test splits.

| Metric | Training-data ratio 0.5 | Training-data ratio 0.8 |
| --- | --- | --- |
| Rare-ordering active rate | $0.5949 \pm 0.231$ | $0.3746 \pm 0.2076$ |
| Compatible SG union size | $14.16 \pm 1.41$ | $12.97 \pm 1.08$ |
| Top-5 compatible candidate count | $4.745 \pm 0.118$ | $4.534 \pm 0.190$ |
| All-Top-5-compatible rate | $0.789 \pm 0.090$ | $0.685 \pm 0.116$ |
| Actionable-case rate | $0.159 \pm 0.06$ | $0.257 \pm 0.094$ |
| Minor-SG actionable-case rate | $0.117 \pm 0.069$ | $0.197 \pm 0.110$ |

As summarized in **Table S6**, the rare-ordering active rate decreased as the training-data ratio increased. Consistently, the compatible SG union size also decreased, indicating that ordering information became less diffuse and more selective in the high-data regime. In addition, both the Top-5 compatible candidate count and the all-Top-5-compatible rate decreased, suggesting that the retrieved Top-5 candidate set more often contained both compatible and incompatible candidates. Under these conditions, ordering information can provide a more discriminative structural prior.

This trend was also reflected in the actionable-case rate. The overall actionable-case rate increased with a higher training-data ratio, and a similar pattern was observed for the minor-SG subset. These results suggest that ordering information does not become intrinsically more important simply because more training data are available. Rather, in the high-data regime, the retrieved candidate-set structure more frequently satisfies the conditions under which ordering information can act as an effective soft structural prior.

Overall, this analysis explains why removing ordering information had a limited and mixed effect at a lower training-data ratio but led to a clearer performance decrease at a higher training-data ratio. At a lower training-data ratio, the active compatible SG set was broader, and most retrieved Top-5 candidates were already ordering-compatible. In this case, ordering information could retain the correct SG but had limited ability to discriminate among candidate SGs. At a higher training-data ratio, the compatible SG set became narrower, and the retrieved Top-5 candidate set more often contained both compatible and incompatible candidates. Under these conditions, ordering information could more effectively serve as a soft structural prior for adjusting boundary candidates in the final SG prediction.

## 7. Quantitative-indicator analysis of minor-SG correction cases

To examine how DyRIS corrected retrieval-only Top-1 errors for minor SGs, we analyzed cases in which the ground-truth label was a minor SG, the retrieval-only Top-1 prediction was incorrect, and the final DyRIS Top-1 prediction was correct. In these cases, the DyRIS-selected Top-1 SG corresponds to the ground-truth SG. Therefore, we compared the ground-truth SG with the retrieval-only Top-1 candidate using the quantitative indicators provided in the prompt.

**Table S7.** Metric-wise comparison between the ground-truth SG and the retrieval-only Top-1 candidate in minor-SG correction cases. The analysis was performed for samples whose ground-truth labels were minor SGs, whose retrieval-only Top-1 predictions were incorrect, and whose final DyRIS Top-1 predictions were correct. For each quantitative indicator, the table reports the number of cases in which either the ground-truth SG or the retrieval-only Top-1 candidate showed the better indicator value, along with the number of ties.

| | **Training-data ratio = 0.5** | | | **Training-data ratio = 0.8** | | |
|---|---|---|---|---|---|---|
| **Metric** | **Ground-truth SG** | **Retrieval Top1 wins** | **Tie** | **Ground-truth SG** | **Retrieval Top1 wins** | **Tie** |
| Combined score | 135 | 24 | 1 | 100 | 24 | 0 |
| Global fit | 140 | 20 | 0 | 117 | 7 | 0 |
| $z_{abs_mean}(s)$ | 145 | 15 | 0 | 117 | 7 | 0 |
| $z_{max_abs}(s)$ | 134 | 26 | 0 | 105 | 19 | 0 |
| $z_{best_count}(s)$ | 121 | 16 | 23 | 92 | 16 | 16 |

Across both training-data ratios, the ground-truth SG more frequently showed better indicator values than the retrieval-only Top-1 candidate for all evaluated quantitative indicators. These results support the interpretation that DyRIS improves Top-1 prediction for minor SGs by integrating quantitative evidence rather than simply following the retrieval order.

## 8. DyRIS–PyCaret hybrid heuristic

The heuristic rules used for the DyRIS–PyCaret hybrid approach are summarized below.

**First,** the hybrid approach applies a DyRIS-aware Top-1 preservation rule. If the Top-1 candidate predicted by DyRIS appears within the Top-2 candidates predicted by the PyCaret-based model, the DyRIS Top-1 candidate is selected as the final Top-1 candidate. This rule preserves DyRIS-supported predictions when they are also considered highly plausible by the data-driven model.

**Second,** if the DyRIS-aware preservation rule is not activated, the hybrid approach applies agreement-first Top-1 selection. The Top-3 candidates predicted by DyRIS and the PyCaret-based model are compared. If one or more overlapping SG candidates are present, the overlapping candidate with the smallest rank-sum across the two models is selected as the initial Top-1 candidate. The rank-sum is defined as the sum of the candidate's rank in the DyRIS Top-3 list and its rank in the PyCaret Top-3 list. If no overlapping candidate exists, the candidate with the smallest rank-sum-like score across the union of both candidate lists is selected.

**Third,** a symmetric majority-class bias correction is applied to SG 14 and SG 225. Because SG 14 and SG 225 are highly frequent in the training data, a major-SG prediction supported by only one model can reflect majority-class bias. Therefore, if the provisional Top-1 candidate is SG 14 or SG 225 and is not supported by both models, the hybrid approach considers replacing it with the strongest candidate from the other model. The replacement is accepted when the alternative candidate is either a non-major SG or is also supported by both models. In contrast, if SG 14 or SG 225 is supported by both DyRIS and PyCaret, it is retained because both the rule-guided inference and the data-driven classifier support the same majority-class prediction.

**Finally,** after the final Top-1 candidate is fixed, the remaining Top-3 positions are filled from the union of DyRIS and PyCaret candidates. Duplicate SG labels are removed. The remaining candidates are ranked by a combined priority score that favors lower rank-sum across the two models, candidates supported by both models, and candidates that appear at higher rank in either model. This process continues until three unique SG candidates are obtained. The final output of the hybrid approach is a ranked Top-3 SG list.